\documentclass[publish]{smj}

\usepackage[greek,UKenglish]{babel}

\usepackage[autostyle=false, style=english]{csquotes}
\MakeOuterQuote{"}
\usepackage{ragged2e} 
\usepackage{tabularx}
\usepackage{booktabs}
\usepackage{threeparttable}
\usepackage{xcolor}
\usepackage{adjustbox}
\usepackage{algpseudocode}
\usepackage{algorithm}
\usepackage{siunitx}

\DeclareMathOperator{\diag}{diag}

\DeclareMathOperator{\Mult}{Mult}
\DeclareMathOperator{\InvGamma}{Inv \, Gamma}

\DeclareMathOperator{\softmax}{softmax}

\Author{Schyan Zafar\Affil{1} \and Geoff K. Nicholls\Affil{1}}
\AuthorRunning{Schyan Zafar \and Geoff K. Nicholls}
  
\Affiliations{

\item Department of Statistics,       
      University of Oxford,
      Oxford,
      UK

}   

\CorrAddress{Schyan Zafar, 
             Department of Statistics, \\
             24--29 St Giles',
             Oxford, 
             OX1 3LB, 
             United Kingdom}
\CorrEmail{sczafar@stats.ox.ac.uk}
\CorrPhone{(+44)\;1865\;272860}

\Title{An Embedded Diachronic Sense Change Model with a Case Study from Ancient Greek}
\TitleRunning{An Embedded Diachronic Sense Change Model}

\Abstract{
Word meanings change over time, and word \textit{senses} evolve, emerge or die out in the process. For ancient languages, where the corpora are often small and sparse, modelling such changes accurately proves challenging, and quantifying uncertainty in sense-change estimates consequently becomes important. GASC (Genre-Aware Semantic Change) and DiSC (Diachronic Sense Change) are existing generative models that have been used to analyse sense change for target words from an ancient Greek text corpus, using unsupervised learning without the help of any pre-training. These models represent the senses of a given target word such as ``kosmos'' (meaning decoration, order or world) as distributions over context words, and sense prevalence as a distribution over senses. The models are fitted using Markov Chain Monte Carlo (MCMC) methods to measure temporal changes in these representations. This paper introduces EDiSC, an Embedded DiSC model, which combines word embeddings with DiSC to provide superior model performance. It is shown empirically that EDiSC offers improved predictive accuracy, ground-truth recovery and uncertainty quantification, as well as better sampling efficiency and scalability properties with MCMC methods. The challenges of fitting these models are also discussed.}


\begin{document}

\maketitle


%

\section{Introduction}

Languages evolve with time, and there are many facets to linguistic change that are of considerable interest to researchers across a wide range of academic disciplines. One such facet is diachronic lexical semantics, which is the study of how the meanings of a word change with time. It is a complex phenomenon that can be influenced by a variety of factors, including social, cultural and technological changes. Computational methods for diachronic semantic change analysis have become increasingly popular in recent years, as they offer a way to study corpora of text data in a systematic and objective manner.

In this paper, the particular problem within diachronic lexical semantics that we are interested in is that of modelling polysemy, i.e. multiple meanings for a word, and homography (a subset of homonymy), i.e. words with the same spelling but different meanings. An example of a polyseme is the word ``mouse'' (meaning a rodent or a computer pointing device), whereas ``bear'' (the animal) and ``bear'' (to carry) are examples of homographs. (We do not distinguish between polysemy and homography in this paper.) We are interested in learning and modelling the different meanings or \textit{senses} of given polysemous/homographic target words over time, and quantifying the uncertainty in these sense-change estimates, from text data that does not have sense labels. The unsupervised nature of this task makes it particularly challenging for small and sparse text datasets.

\citet{perrone-etal-2019-gasc}, building on the framework of the Bayesian Sense ChANge (SCAN) model of \citet{frermann2016bayesian}, introduced a model called Genre-Aware Semantic Change (GASC), and applied it to dynamically model the senses of selected target words from an ancient Greek text corpus. In this framework, distinct senses of a target word are represented as distinct distributions over context words, sense prevalence is represented as a distribution over target-word senses, and these distributions are allowed to evolve with time. \citet{DiSC_https://doi.org/10.1111/rssc.12591} further built on this framework, and modelled sense and time as additive effects in their Diachronic Sense Change (DiSC) model, which offered much improved model performance. They quantified uncertainty in the sense-change estimates using credible sets for the evolving distributions, and showed that these agreed well with those given by expert annotation for their test cases.

Quantification of uncertainty in sense-change estimates is an under-explored area within the field, yet important when working with small datasets. (Typically, any dataset with under 40 million tokens is considered `small', whereas our ancient Greek data has around 10 million tokens.) It is particularly important for historic corpora where training data are limited and sparse (i.e. with a large proportion of infrequently used words), and inferences drawn as point estimates are therefore unreliable. Fitting models to these data is no easy task, and requires careful statistical modelling to draw accurate and meaningful inferences. The difficulty is compounded since SCAN, GASC and DiSC are all fitted to subsets of text \textit{snippets} (i.e. fixed-length windows of context words surrounding the target), thus ignoring the information contained in the wider text corpora outside the snippets, which affects the quality of fit and the inferences drawn. 

In this paper, we develop EDiSC, an Embedded Diachronic Sense Change model, which extends the DiSC model by combining it with word embeddings, whereby context words are represented as vectors in an embedding space. 
This has two main advantages. Firstly, embeddings exploit the wider text corpora to capture useful semantic information about the context words, which is otherwise lost if we focus only on the context of a given target word. This feature of EDiSC leads to improved predictive accuracy, ground-truth recovery and uncertainty quantification. We demonstrate this on challenging test cases from ancient Greek, which is the main focus of this paper, as well as an easier test case from English. Secondly, the dimension of the embedding space is lower than the vocabulary size, and is typically held constant even against an increasing vocabulary size. This results in more efficient Monte Carlo sampling and scalability properties. We demonstrate this via experiments on both real and synthetic data.

The main novelty of this paper is to bring together two previously separate approaches for modelling sense change, namely topic-based models and embedding-based models, to improve model performance. This was not straightforward, as careful statistical modelling and consideration of the model structure are required to make things work. Furthermore, compared to \citet{DiSC_https://doi.org/10.1111/rssc.12591}, we treat model selection explicitly, as well as giving a more thorough treatment of model-fitting methods and convergence issues. We also consider additional ancient Greek test cases, in particular one where accurate sense-change analysis was not achievable using DiSC, but is now possible.

The rest of this paper is organised as follows. In Section~\ref{sec:data_and_problem} we describe the datasets used and our modelling/inference objectives. In Section~\ref{sec:related_work} we discuss related work and where our paper fits within the wider literature. In Section~\ref{sec:model} we describe our new model and how it relates to existing models. We also discuss the embeddings used and inference for our model. In Section~\ref{sec:results} we show the results of applying the models to our test cases, and assess model performance on predictive accuracy and true-model recovery. We also discuss model selection issues, and sampling efficiency and scalability. Section~\ref{sec:discussion} concludes with a brief discussion. The Appendix contains a discussion on hyperparameter settings, further results not included in the main body, and some other technical details.

\section{Data and problem setting} \label{sec:data_and_problem}

Whilst the models discussed in this paper could be applied quite widely to gain useful insights on potential target words with sense change, in this paper we are primarily interested in modelling sense change for three target words from the Diorisis Ancient Greek Corpus \citep{TheDiorisisAncientGreekCorpus}: ``kosmos'', ``mus'' and ``harmonia''. We work with this corpus as it is small and sparse, and therefore challenging for existing methods. We choose these particular test cases since we have the ground truth readily available for these target words, which we can use to accurately assess model performance. Otherwise, we would be limited to qualitative measures of model performance, which are not as convincing. (The ground truth would, of course, not be available for a `real-use' case.) Fitting the models for these test cases is particularly challenging; so, for demonstration purposes, we also use a simple test case, ``bank'', from the Corpus of Historical American English (COHA, \citealt{davies2010corpus}), where fitting the models is much easier.

\begin{table*}[!t]
\caption{Example text snippets for target word ``\textcolor{red}{bank}'' taken from \citet[Table~1]{DiSC_https://doi.org/10.1111/rssc.12591}. The words are lemmatised, and stopwords in \textcolor{blue}{blue} and infrequent words in \textcolor{orange}{orange} are dropped to get the data $W$.} \label{tab:bank_snippets}
\centering
\begin{tabularx}{\textwidth}{X}
\toprule
``... \textcolor{orange}{China}\textcolor{darkgray}{.} \textcolor{blue}{The} \textcolor{orange}{Yellow} River \textcolor{blue}{had} burst \textcolor{blue}{its} \textcolor{red}{banks}\textcolor{darkgray}{,} submerging vast areas \textcolor{blue}{of} \textcolor{orange}{farmland}\textcolor{darkgray}{,} washing \textcolor{blue}{away} ...''\\
\midrule
``... \textcolor{blue}{to} examine \textcolor{blue}{whether} institutions \textcolor{blue}{like the World} \textcolor{red}{Bank} \textcolor{blue}{and the} International Monetary Fund \textcolor{blue}{needed} \textcolor{orange}{restructuring} ...''\\
\midrule
``... subject \textcolor{blue}{of} \textcolor{orange}{continuing} specie payments\textcolor{darkgray}{.} \textcolor{blue}{Though the} \textcolor{red}{Bank} \textcolor{blue}{of the United States had previously} determined ...''\\
\bottomrule
\end{tabularx}
\end{table*}

COHA consists of 400+ million tokens, from which \citet{DiSC_https://doi.org/10.1111/rssc.12591} randomly select 3,685 snippets of 14 \textit{lemmatised} context words around the target word ``bank'', and annotate them with the sense of riverbank or financial institution. (A \textit{lemma} is the root form of the word; so for example ``branch'' is the lemmatised form of ``branching''.) The annotation is done to provide ground truth for testing, and is not used in the analysis. Some example snippets are shown in Table~\ref{tab:bank_snippets}. Of these sense-labelled snippets, 3,525 snippets of the type \textit{collocates} are used for evaluation, i.e. the snippets where the correct target-word sense could be identified by the reader from context alone. Stopwords and infrequently used words are dropped, leaving a vocabulary of 973 words, and punctuation and sentence boundaries are ignored. Time is divided into ten discrete and contiguous 20-year periods from 1810 to 2010, and a single (combined) genre is used. The choices of snippet length, time discretisation and genre covariate are subjective. These could be informed by domain knowledge or exploratory analysis, or pre-set based on the quantities of interest. We use the same choices as in previous work since that is not our focus, and they are in any case well informed. 

For the ancient Greek data, which is a much smaller corpus with around 10 million tokens, expert sense-annotation is provided by \citet{vatri_lahteenoja_mcgillivray_2019} for our three target words, with an accompanying explanation by \citet[Section~4]{10.1093/llc/fqz036}. Each target word has three true senses: ``kosmos'' (decoration, order, world), ``mus'' (mouse, muscle, mussel), ``harmonia'' (abstract, concrete, musical). 
Following \citet{perrone-etal-2019-gasc}, time periods are discrete and contiguous centuries from 800~BC to 400~AD, and the categorical genre covariates are narrative and non-narrative for the ``kosmos'' data, and technical and non-technical for the ``mus'' and ``harmonia'' data.
We use snippets of 14 lemmatised context words in line with \citet{DiSC_https://doi.org/10.1111/rssc.12591}. However, in contrast to their setup, we filter the vocabulary based on a minimum count of 10 occurrences in the entire corpus. This filtering is used to reduce noise from rare words. In order to avoid filtering words that are rare in the target-word context, but common outside that context, we base these counts on frequency in the corpus rather than just the snippets. Further, we filter out a smaller list of stopwords than that used by \citet{DiSC_https://doi.org/10.1111/rssc.12591}, preferring to retain some potentially noisy context words rather than lose some context words that may be semantically important. 
(Previous work identified stopwords using part-of-speech tags, as well as three lists: (a) \citet{greek_stopwords}, and (b) \texttt{misc} and (c) \texttt{stopwords-iso} from the R package \texttt{Stopwords} \citep{stopwords2020}. We continue to use part-of-speech tags as before, but only use the last of the three lists.)
Therefore, whilst the Greek datasets used are comparable, they are not identical; and we report all results on the reprocessed data. Also, \citet{DiSC_https://doi.org/10.1111/rssc.12591} only analysed ``kosmos'', whereas we now fit the models to all three datasets. 

Table~\ref{tab:data_summary} summarises the four datasets and some notation that we will use in this paper. Note that the numbers differ from \citet[Table~3]{10.1093/llc/fqz036} because of the slightly different approaches used to extract the snippets. The data $W$ for a given target word consists of $D$ snippets of length $L$ containing context words $w_{d,i}, d \in 1:D, i \in 1:L$, sampled from the vocabulary $1:V$. A subset of $D'$ snippets is of the type collocates. A context position may be empty in the filtered data if a stopword or infrequent word has been dropped; so the bag of words retained in snippet $d$, denoted $W_d$, has variable size $L_d$. The target word itself is excluded. Snippet $d$ has deterministic mappings to genre $\gamma_d \in 1:G$ and time $\tau_d \in 1:T$, which are known from the text that generated the snippet. The target-word sense $z_d \in 1:K$ for snippet $d$ is in general unknown. We use $K'$ to refer to the number of true target-word senses, whereas we fit the models using $K$ senses, which may be different to $K'$. The choice of $K$ is discussed in Section~\ref{sec:model_selection}. 


\begin{table}[t]
\caption{Data summary} \label{tab:data_summary}
\centering
\begin{tabular}{l l r r r r}
\toprule
Data & ($W$)            & bank & kosmos & mus & harmonia \\
\midrule
Snippets & ($D$)        & 3,685 & 1,469 & 214 &   653 \\
Collocates & ($D'$)     & 3,525 & 1,144 & 118 &   451 \\
Vocabulary size & ($V$) &   973 & 2,904 & 899 & 1,607 \\
Snippet length & ($L$)  &    14 &    14 &  14 &    14 \\
True senses & ($K'$)    &     2 &     3 &   3 &     3 \\
Model senses & ($K$)    &     2 &     4 &   3 &     4 \\
Text genres & ($G$)     &     1 &     2 &   2 &     2 \\
Time periods & ($T$)    &    10 &     9 &   9 &    12 \\
\bottomrule
\end{tabular}
\end{table}

In practice, given a text corpus and target word of interest, the workflow to extract data $W$ might go as follows: identify target-word occurrences in corpus $\rightarrow$ set snippet length $L$ $\rightarrow$ extract snippets and meta-data (time and genre) $\rightarrow$ lemmatise snippets $\rightarrow$ set criteria for stopwords and minimum frequency $\rightarrow$ filter these words from snippets $\rightarrow$ discretise time and set genre covariates to get data $W$. Given these data, our goal is to dynamically model the target-word senses and sense prevalence where, separately for each target word, we define sense $k$ at time $t$ as the distribution $\tilde{\psi}^{k,t} = \tilde{\psi}^{k,t}_{1:V}$ over context words $1:V$, and we define sense prevalence for genre $g$ at time $t$ as the distribution $\tilde{\phi}^{g,t} = \tilde{\phi}^{g,t}_{1:K}$ over senses $1:K$. We would like to infer the context-word probabilities $\tilde{\psi}^{k,t}_v = p(w_{d,i}=v | z_d=k, \tau_d=t)$ for each $(v,k,t)$ word-sense-time triple, the sense-prevalence probabilities $\tilde{\phi}^{g,t}_k = p(z_d=k | \gamma_d=g, \tau_d=t)$ for each $(k,g,t)$ sense-genre-time triple, and quantify the uncertainty therein. Then, $\tilde{\psi}$ and $\tilde{\phi}$ would together encapsulate the diachronic sense change over time~$t$.


\section{Related work} \label{sec:related_work}

The field of computational lexical semantic change is quite rich and varied. Approaches include topic-based models (e.g. \citealt{frermann2016bayesian, perrone-etal-2019-gasc, DiSC_https://doi.org/10.1111/rssc.12591}), on which our current work builds, as well as graph-based models (e.g. \citealt{2014arXiv1405.4392M, mitra2015automatic, tahmasebi2017finding}) and embedding-based models (e.g. \citealt{Kulkarni:2015:SSD:2736277.2741627, 2016arXiv160509096H, Rudolph:2018:DEL:3178876.3185999, 2019arXiv190601688D, 9581306}). Comprehensive surveys tracing the history of the subject up to 2021 are given by \citet{tang2018state}, \citet{kutuzov-etal-2018-diachronic} and \citet{tahmasebi2021survey}.

In recent years, embedding-based methods have dominated the landscape of natural language processing (NLP). Word embeddings are representations of words in low-dimensional real vector spaces, and embedding models produce these mappings whilst capturing the words' semantic and (sometimes) syntactic relationships. Popular traditional word embedding models include Google's Word2vec \citep{2013arXiv1301.3781M, NIPS2013_5021, mikolov-etal-2013-linguistic}, Stanford's GloVe \citep{pennington2014glove} and Facebook's FastText \citep{10.1162/tacl_a_00051}, but there are many others, including variants of these models. Traditional word embeddings are learnt based on patterns of word co-occurrences in text corpora. These are typically context-independent, i.e. the embedding for a word is the same regardless of the context in which it is used. This is a limitation for diachronic sense change modelling, since many words have multiple senses which traditional embeddings fail to capture.

Contextualised word embedding models have been developed in recent years to overcome this limitation by attempting to learn word representations that capture word meanings in their given contexts. Some early contextualised embedding models such as Context2vec \citep{melamud2016context2vec} and ELMo (Embeddings from Language Model, \citealt{peters-etal-2018-deep}) are based on recurrent neural network architectures, whereas most modern contextualised embedding models are based on transformer architectures, including the popular GPT (Generative Pre-trained Transformer, \citealt{radford2019language}) and BERT (Bidirectional Encoder Representations from Transformers, \citealt{devlin-etal-2019-bert}). Recent surveys discussing modern embedding models, including many variants of the ones mentioned here, are given by \citet{10.1145/3434237} and \citet{10.1162/coli_a_00474}. 

Word embeddings can be used for a number of NLP tasks, including tasks within computational semantics. Contextualised word embeddings in particular can be used for diachronic sense change modelling, and \citet{2023arXiv230401666M} survey the recent advances in this area. Most modern methods are data-intensive, and tend to rely on pre-trained models learnt using huge amounts of data. Indeed, virtually all of the approaches for modelling semantic change summarised in \citet[Tables~3-4]{2023arXiv230401666M} rely on some form of pre-training, fine-tuning or domain-adaptation. This restricts the usability of these methods for analysing small, sparse or historic corpora, such as the ancient Greek data used in this paper, for which training data are limited, and pre-trained models are not available. 

Word sense disambiguation (WSD) is the related task of identifying the correct sense of a word in a given context. Recent approaches for WSD tend to be either knowledge-based or supervised, utilising sense inventories or sense-annotated corpora, and typically rely on pre-trained language models \citep{bevilacqua2021recent}. This again restricts the usage of WSD approaches for our purpose. 

Moreover, the goal of WSD is not diachronic sense change modelling, but rather sense induction from context. The usages of a target word may be clustered according to the induced senses, which could then be used for drawing inferences about diachronic sense change, but a clustering approach is not conducive to interpretability or quantification of uncertainty. Graph-based approaches have a similar drawback. On the other hand, generative models designed for the purpose of diachronic sense change modelling have parameters with physical interpretations, and are natural in the context of Bayesian measures of uncertainty.

Probabilistic topic models are generative bag-of-words models that are widely used to infer themes or \textit{topics} from a collection of documents \citep{blei2012probabilistic}. Latent Dirichlet Allocation (LDA, \citealt{blei2003latent}), one of the best-known topic models, is a generative model that represents topics as distributions over words, and documents as mixtures over topics. The dynamic topic model (DTM, \citealt{Blei:2006:DTM:1143844.1143859}) extends LDA to model each topic as a time series, whereas the embedded topic model (ETM, \citealt{10.1162/tacl_a_00325}) extends LDA to incorporate word embeddings within the model. The ETM outperforms LDA in terms of topic quality and predictive performance since it benefits from the extra semantic information captured within the word embeddings that is not present in LDA. Finally, the dynamic embedded topic model (D-ETM, \citealt{dieng2019dynamic}) brings together the DTM and the ETM by adding a time dimension to the ETM. The D-ETM similarly outperforms the DTM, while requiring less time to fit. New variants of topic models continue to emerge \citep{churchill2022evolution}, showing their continued relevance and importance.

The DTM was adapted by \citet{frermann2016bayesian} to capture the evolving \textit{senses} of a given target word in the SCAN model. A snippet under SCAN is the context surrounding an instance of the target word, and may be compared to a document under the DTM. The distinction is that, whilst a document under the DTM is a mixture over multiple topics, each snippet under SCAN has only one sense. The sense prevalence was allowed to vary according to the text genre by \citet{perrone-etal-2019-gasc} in the GASC model. Each sense in the SCAN and GASC models has an independent time-evolution, whereas \citet{DiSC_https://doi.org/10.1111/rssc.12591} imposed an additive structure between the sense-effect and the time-effect in the DiSC model. Therefore, under DiSC, differentiated target-word senses have a common time-evolution, which drastically reduces the dimension of the parameter space. DiSC captures the sense-dependence of snippets better than SCAN/GASC, and results in more accurate sense induction and diachronic sense change modelling.

The relationship between DiSC and the EDiSC model we introduce in this paper is analogous to the relationship between the DTM and the D-ETM. In both cases, the latter extends the former through incorporating word embeddings within the model. Indeed, EDiSC has been inspired by the D-ETM, and offers similar benefits over DiSC as the D-ETM does over DTM.

\section{Model and inference} \label{sec:model}

We first describe DiSC, highlighting its connection to GASC and SCAN, before introducing our new EDiSC model. We then discuss the embeddings used and the inference methods for these models. Prior elicitation with respect to the hyperparameters is discussed in Appendix~\ref{sec:hyperparameters}.

\subsection{Background}

DiSC is a generative bag-of-words model for the context words surrounding a given target word, comprising a prior model and an observation model.


Under the DiSC observation model, to generate snippet $d$, we first sample the sense $z_{d} | \tilde{\phi}^{\gamma_d,\tau_d} \sim \Mult ( \tilde{\phi}^{\gamma_d,\tau_d} )$ and then independently sample the words $w_{d,i} | z_d, \tilde{\psi}^{z_d,\tau_d} \sim \Mult ( \tilde{\psi}^{z_d,\tau_d} )$ for each context position $i$ in the snippet. The context positions occupied by words in each snippet $d$ are a subset $\{ i_1 , \dots, i_{L_d} \}$ of size $L_d$ drawn from the set $\{1,\dots,L\}$, where $L_d$ is the number of words retained in snippet $d$ after filtering out stopwords and infrequent words. The order of context words is irrelevant due to the bag-of-words assumption. The authors of DiSC treat stopwords and infrequent words explicitly, whereas the authors of GASC and SCAN do not. This is a difference in how context words are registered within the snippets rather than a difference in the models. The observation models are identical for DiSC, GASC and SCAN. 


Under the DiSC, GASC and SCAN prior models, $\tilde{\phi}^{g,t}$ and $\tilde{\psi}^{k,t}$ are defined as softmax transforms of real-valued vectors $\phi^{g,t}$ and $\psi^{k,t}$ respectively, that is
\begin{equation}
    \tilde{\phi}^{g,t} = \frac {\exp (\phi^{g,t})} {\sum_{k=1}^K \exp (\phi_{k}^{g,t})}
    \quad \text{ and } \quad 
    \tilde{\psi}^{k,t} = \frac {\exp (\psi^{k,t})} {\sum_{v=1}^V \exp (\psi_{v}^{k,t})} \text{.} \label{eq:softmax}
\end{equation}
Under the DiSC prior model, for each genre $g$, the prior on $\phi^{g,t}$ is an AR(1) time series with stationary distribution $\mathcal{N} \left( 0, \diag \left( \frac{\kappa_{\phi}} {1 - (\alpha_{\phi})^2} \right) \right)$. An additive structure $\psi^{k,t} = \chi^k + \theta^t$ is imposed on $\psi$. A $\mathcal{N} ( 0, \diag(\kappa_{\chi}) )$ prior is placed on $\chi^k$, and an AR(1) prior is placed on $\theta^t$ with stationary distribution $\mathcal{N} \left( 0, \diag \left( \frac{\kappa_{\theta}} {1 - (\alpha_{\theta})^2} \right) \right)$. Prior hyperparameters $\kappa_{\phi}, \kappa_{\theta}, \kappa_{\chi}, \alpha_{\phi}, \alpha_{\theta}$ are fixed, and $|\alpha_{\phi}| < 1, |\alpha_{\theta}| < 1$ to ensure proper priors with mean reversion.

In contrast, GASC and SCAN have no additive structure on $\psi^{k,t}$ and model it as $K$ independent Gaussian time series (without mean-reversion). All time series priors used in GASC and SCAN are improper without a stationary distribution: $\phi^{g,t} | \phi^{g,-t}, \kappa_\phi \sim \mathcal{N} \left( \frac{1}{2} (\phi^{g,t-1} + \phi^{g,t+1}), \kappa_\phi \right)$ and $\psi^{k,t} | \psi^{k,-t}, \kappa_\psi \sim \mathcal{N} \left( \frac{1}{2} (\psi^{k,t-1} + \psi^{k,t+1}), \kappa_\psi \right)$. Also, $\kappa_\phi \sim \InvGamma(a,b)$ in GASC and SCAN. Finally, SCAN has the number of genres $G=1$ fixed, whereas GASC admits $G \geq 1$; otherwise, the two models are identical.

\subsection{Embedded DiSC (EDiSC) model} \label{sec:EDiSC_model}

\begin{algorithm}[!t]
\caption{EDiSC: generative model}
\label{alg:EDiSC_generative_model}
\begin{algorithmic}[1]

\centering
\Statex --------------------- PRIOR MODEL ---------------------

\raggedright
\State get word embeddings matrix $\rho$
\State fix hyperparameters $\kappa_{\phi}, \kappa_{\theta}, \kappa_{\chi}, \kappa_{\varsigma}, \alpha_{\phi}, \alpha_{\theta}$ (with $|\alpha_{\phi}| < 1, |\alpha_{\theta}| < 1$) 

\State draw bias or correction parameter $\varsigma | \kappa_{\varsigma} \sim \mathcal{N} \left( 0, \diag(\kappa_{\varsigma}) \right)$ 

\For {genre $g \in 1:G$}
    \State draw initial sense prevalence parameter $\phi^{g,1} | \kappa_{\phi}, \alpha_{\phi} \sim \mathcal{N} \left( 0, \diag \left( \frac{\kappa_{\phi}} {1 - (\alpha_{\phi})^2} \right) \right)$
    \For {time $t \in 2:T$}    
        \State draw sense prevalence parameter $\phi^{g,t} | \phi^{g,t-1}, \kappa_{\phi}, \alpha_{\phi} \sim \mathcal{N} \left( \alpha_{\phi} \phi^{g,t-1}, \diag ( \kappa_{\phi} ) \right)$
    \EndFor
\EndFor

\State draw initial time embedding $\theta^{1} | \kappa_{\theta}, \alpha_{\theta} \sim \mathcal{N} \left( 0, \diag \left( \frac{\kappa_{\theta}} {1 - (\alpha_{\theta})^2} \right) \right)$
\For {time $t \in 2:T$}
    \State draw time embedding $\theta^{t} | \theta^{t-1}, \kappa_{\theta}, \alpha_{\theta} \sim \mathcal{N} \left( \alpha_{\theta} \theta^{t-1}, \diag ( \kappa_{\theta} ) \right)$
\EndFor

\For {sense $k \in 1:K$}
    \State draw sense embedding $\chi^k | \kappa_{\chi} \sim \mathcal{N} \left( 0, \diag(\kappa_{\chi}) \right)$
    \For {time $t \in 1:T$}
        \State set sense-time embedding $\xi^{k,t} = \chi^k + \theta^t$
        \State set context-word probability parameter $\psi^{k,t} = \rho \xi^{k,t} + \varsigma$
    \EndFor
\EndFor 

\State transform $\phi$ and $\psi$ into probabilities $\tilde{\phi}$ and $\tilde{\psi}$ using softmax \eqref{eq:softmax}

\centering
\Statex --------------- OBSERVATION MODEL --------------- 

\raggedright

\For {snippet $d \in 1:D$ (genre $\gamma_d$, time $\tau_d$, length $L_d$)}
    \State draw sense assignment $z_{d} | \tilde{\phi}^{\gamma_d,\tau_d} \sim \Mult \left( \tilde{\phi}^{\gamma_d,\tau_d}_{1}, \dots, \tilde{\phi}^{\gamma_d,\tau_d}_{K} \right)$
    \For {context position $i \in \{ i_1 , \dots, i_{L_d} \}$}
        \State draw context word $w_{d,i} | z_d, \tilde{\psi}^{z_d,\tau_d} \sim \Mult \left( \tilde{\psi}^{z_d,\tau_d}_{1}, \dots, \tilde{\psi}^{z_d,\tau_d}_{V} \right)$
    \EndFor
\EndFor

\end{algorithmic}
\end{algorithm}

\begin{figure}[!t]
\centering
\includegraphics[width = 0.75\textwidth, keepaspectratio]{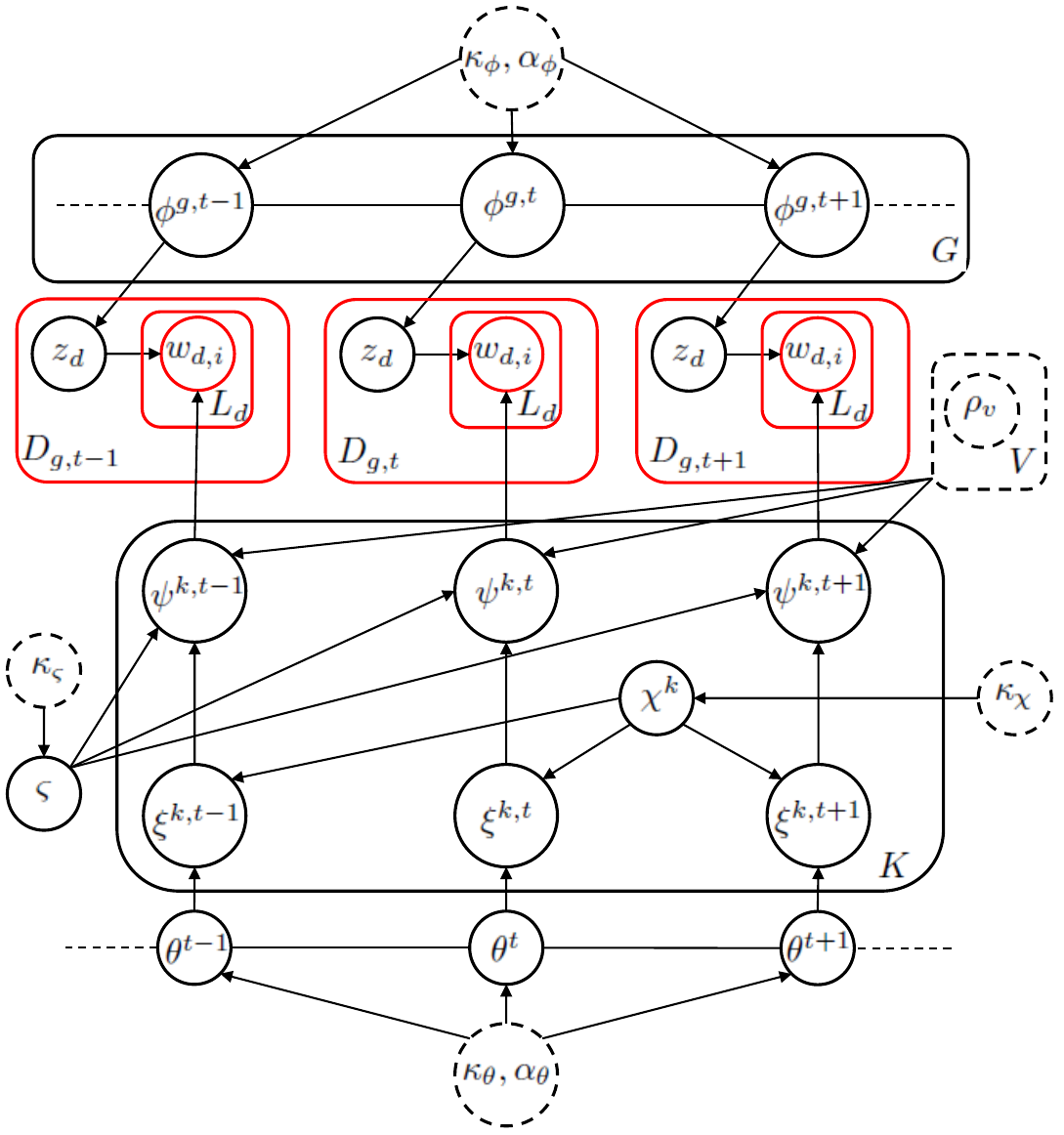}
\caption{EDiSC plate diagram for three time periods. Dashed nodes are constant parameters, solid black nodes are latent variables and solid red nodes are observed variables. $D_{g,t}$ is the number of snippets for genre $g$ at time $t$.}
\label{fig:EDiSC_plate_diagram}
\end{figure}

The EDiSC generative model is given in Algorithm~\ref{alg:EDiSC_generative_model} and a plate diagram is shown in Figure~\ref{fig:EDiSC_plate_diagram}.

The key idea behind EDiSC is to introduce word embeddings into the model. As such, EDiSC has the same observation model as DiSC, as well as the same prior model for $\phi$. However, we now define $\psi^{k,t} = \rho \xi^{k,t} + \varsigma$, where $\rho$ is a $V \times M$ matrix of word embeddings, $\xi^{k,t}$ is an $M$-dimensional sense-time embedding vector for sense $k$ at time $t$, and $\varsigma$ is a $V$-dimensional bias or correction parameter. The matrix $\rho$ has row vectors $\rho_v = (\rho_{v,1}, \dots, \rho_{v,M})$, which are $M$-dimensional word embeddings for words $v \in 1:V$ in the lemmatised vocabulary. The parameter $\xi$ is decomposed as $\xi^{k,t} = \chi^k + \theta^t$, where $\chi^k$ is an $M$-dimensional sense embedding for sense $k$, and $\theta^t$ is an $M$-dimensional time embedding for time $t$. We place a $\mathcal{N} ( 0, \diag(\kappa_{\chi}) )$ prior on $\chi^k$ and an AR(1) prior on $\theta^t$ with stationary distribution $\mathcal{N} \left( 0, \diag \left( \frac{\kappa_{\theta}} {1 - (\alpha_{\theta})^2} \right) \right)$. These priors are functionally the same as in DiSC. However, whilst $\chi^k$ and $\theta^t$ are vectors in a $V$-dimensional space under DiSC, they are now vectors in the $M$-dimensional embedding space under EDiSC. Prior hyperparameters are set using quantiles as discussed in Appendix~\ref{sec:hyperparameters}. 

The bias or correction term $\varsigma$ is used because, due to the coupling induced by the embedding structure, if two words $x,y$ have similar word embeddings $\rho_x, \rho_y$, then in the absence of a correction term the word parameters $\psi^{k,t}_x, \psi^{k,t}_y$ (and hence the probabilities $\tilde{\psi}^{k,t}_x, \tilde{\psi}^{k,t}_y$) would also be very similar. However, we could have context words $x,y$ appearing with different frequencies in the snippets despite the similarity in embeddings. The correction terms $\varsigma_x,\varsigma_y$ serve to decouple the words and allow them to appear with different probabilities. Put another way, if the smaller $M$-dimensional $\xi^{k,t}$ vector is able to capture the variation in the larger $V$-dimensional $\psi^{k,t}$ vector via the product $\rho \xi^{k,t}$, then $\varsigma$ would not be significant; otherwise, $\varsigma$ allows the extra variability required to model $\psi^{k,t}$ accurately.

We briefly mention some other embedded models that we experimented with but discarded. Firstly, we tried $\psi^{k,t} = \rho \chi^k + \theta^t$, using only a sense embedding and not a time embedding, whose predictive performance (as measured by Brier scores) was not as good. Next, we tried $\psi^{k,t} = \rho \xi^{k,t} = \rho (\chi^k + \theta^t)$ without the correction term $\varsigma$, which generally worked well and gave results comparable to those from our chosen final model. However, an issue common to both these alternatives was that the posteriors under these models were more susceptible to multimodality for the sparse Greek data. Hence, using Markov Chain Monte Carlo (MCMC) to find the right mode was less straightforward. 
Finally, we tried $\psi^{k,t} = \rho \xi^{k,t}$ without imposing an additive structure $\xi^{k,t} = \chi^k + \theta^t$, so essentially an embedded version of GASC. The multimodality in the posterior for this model was much worse and, even when the MCMC did converge, the performance was much inferior to our chosen model.

\subsection{Embeddings}

We learn the word embeddings $\rho_v, v \in \{1,\dots,V\},$ using GloVe. Given the model setup in Section~\ref{sec:EDiSC_model}, a traditional embedding model with one vector representation per word is most suitable for use with EDiSC, as we can learn the embeddings independently and plug them into the model. A contextualised embedding model would not be straightforward to use here, although it would be interesting to investigate how it could work within our framework. However, that is beyond the scope of this paper.

Whilst we deliberately choose a popular traditional embedding model, there is no particular reason for choosing GloVe over Word2vec or FastText, and they should all give similar results in our problem setting. We favour an accessible and universally applicable model over a corpus-specific and/or task-specific one (e.g. \citealt{rodda2019a}) for consistency between our English and ancient Greek test cases, and because the code to implement GloVe is readily available.

The R code to implement GloVe was adapted from \citet{GloVe_code}. We learnt the embeddings using the settings in \citet{pennington2014glove}: $x_\text{max}=100$, $\alpha = 3/4$, initial learning rate of 0.05, convergence tolerance of 0.01, context window of 10 words on either side, and adding together the `in' and `out' vectors. We pruned the vocabulary based on a minimum count of 10 in the entire corpus in order to reduce noise from rare words. We filtered out stopwords and lemmatised the words before learning the embeddings, thus tailoring the model to our problem setting by sacrificing syntactic information in favour of semantic. 

In general, there is no universally optimal method for choosing the embedding dimension~$M$. The choice is usually specific to the corpus and/or task. Even then, it is not an exact science and requires judgement. Clearly, a lower dimension would have benefits in terms of computational cost and memory requirements, whereas a larger dimension may capture more semantic information but also introduce noise via overfitting. \citet[Section~4.4]{pennington2014glove} suggest that there are ``diminishing returns for vectors larger than about 200 dimensions'', and \citet{yin2018dimensionality} suggest a method for choosing $M$ based on minimising a loss function. As a rule of thumb, between 50 and 300 is considered an appropriate range \citep{patel-bhattacharyya-2017-towards}. We fit our models for $M$ equal 50, 100, 200 and 300, and report the results for these choices in Section~\ref{sec:predictive_accuracy} below. We also include a brief discussion on how we select $M$ out of these choices in Section~\ref{sec:model_selection}.

\subsection{Inference} \label{sec:inference}

The parameters of interest are the probability arrays $\tilde{\phi}$ and $\tilde{\psi}$, but it is more convenient to target $\phi,\theta,\chi,\varsigma$ given the snippet data $W$. The posterior for EDiSC is defined by 
\begin{equation} \label{eq:EDiSC_posterior}
    \pi(\phi,\theta,\chi,\varsigma | W) \propto \pi(\phi) \pi(\theta) \pi(\chi) \pi(\varsigma) p(W | \phi, \psi) \text{,}
\end{equation}
where the likelihood 
\begin{equation} \label{eq:likelihood}
    p(W | \phi, \psi) = \prod_{d=1}^D \sum_{k=1}^K p(z_d=k | \phi) p(W_d | z_d=k, \psi) = \prod_{d=1}^D \sum_{k=1}^K \tilde{\phi}_k^{\gamma_d,\tau_d} \prod_{w \in W_d} \tilde{\psi}_{w}^{k,\tau_d}
\end{equation}
remains unchanged compared to DiSC in terms of $\phi$ and $\psi$. The likelihood \eqref{eq:likelihood} is obtained by marginalising $p(W,z|\phi,\psi)$ over the unknown sense labels $z=(z_1,\dots,z_D)$. This leaves us with a posterior defined over continuous variables only, which is convenient for variational inference and gradient-based Monte Carlo. The `observable' parameters $\tilde{\phi}$ and $\tilde{\psi}$ are identifiable (up to label switching) whereas the logit-scale parameters $\phi,\theta,\chi,\varsigma$ are not. However, non-identifiability at that level is not a concern, since we only care about the interpretable probability arrays, and non-identifiability does not cause any convergence problems in our experiments. 

The posterior \eqref{eq:EDiSC_posterior} is quite challenging to sample. This is because of ridge structures and multimodality in the posterior, especially for the sparse ancient Greek data, but also to some extent for the less sparse ``bank'' data. Before drawing any inferences, it is important to ensure that any method targeting the posterior has converged, that is, different starting configurations and random-number-generator seeds result in the same posterior distribution. Variational methods targeting \eqref{eq:EDiSC_posterior} are highly sensitive to the starting configuration for the optimisation, and therefore fail this test. Using Stan's Automatic Differentiation Variational Inference (ADVI, \citealt{2015arXiv150603431K}) for instance, we are unable to obtain consistent posteriors for most choices of $M$ for all of our test cases. In any case, variational methods typically target local optima: even when they are adequate for predictive inference (targeting the posterior predictive for $W$), quantification of uncertainty in our target parameters $\tilde\phi$ and $\tilde\psi$ is poor.

MCMC is the method of choice when sampling from \eqref{eq:EDiSC_posterior}, with gradient-based MCMC as described in \citet[Appendix~C]{DiSC_https://doi.org/10.1111/rssc.12591} working particularly well, since this provides better coverage of the posterior space and more accurate quantification of uncertainty. We implement all our samplers in the R programming language \citep{R_citation} and the scripts are available online. We use Metropolis-Adjusted Langevin Algorithm (MALA, \citealt{10.2307/3318418}, \citealt{doi:10.1111/1467-9868.00123}) and Hamiltonian Monte Carlo (HMC, \citealt{1987PhLB..195..216D}, \citealt{2012arXiv1206.1901N}, \citealt{beskos2013}), as well as the No-U-Turn sampler (NUTS, \citealt{hoffman2014no}) from the Stan software \citep{Stan_citation, RStan_citation}. 

In our implementations of MALA and HMC, described in more detail in Appendix~\ref{sec:samplers}, we use analytically derived gradients; and we target \eqref{eq:EDiSC_posterior} in a Metropolis-within-Gibbs fashion, alternately sampling each variable given the others. Stan, on the other hand, uses automatic numerical differentiation and targets the entire posterior at once. The choice of sampler is not important in any converged MCMC run, since all samplers should converge to the same posterior. However, due to the ridge structures in \eqref{eq:EDiSC_posterior}, the samplers may sometimes get stuck in a metastable state and fail to converge. (A metastable state is a region of high density but low total probability mass, separated from the rest of the posterior by regions of low density.) Our HMC sampler generally gives the most consistent performance in this respect. Some of the convergence issues experienced in fitting these models to our test cases are discussed in Appendix~\ref{sec:convergence}. In Section~\ref{sec:results} below, we only report results from converged HMC runs unless otherwise stated.

\section{Application and results} \label{sec:results}

We first consider the models' predictive accuracy on held-out true sense labels. We then discuss model selection issues with respect to the choice of $K$ and $M$. Next, we assess the inferred sense-prevalence evolution of our target words against the ground truth. Finally, we analyse the sampling efficiency and scalability properties of EDiSC vs. DiSC using MCMC methods.

\subsection{Predictive accuracy} \label{sec:predictive_accuracy}

We use the held-out true sense labels $o_d \in \{1',\dots,K'\}$, where $d \in \{1',\dots,D'\}$ are the indices for the type-collocates snippets (cf. Section~\ref{sec:data_and_problem}), to assess the models' predictive performance. We quantify this using the Brier score 
\begin{equation*}
    \text{BS} = \frac{1}{D'} \sum_{d=1'}^{D'} \sum_{k=1'}^{K'} \left( \hat{p}(z_d=k) - \mathbb{I} (o_d=k) \right)^2 \text{,}
\end{equation*}
a proper scoring rule for multi-category probabilistic predictions $\hat{p}(z_d=k)$, ranging from 0 (best) to 2 (worst). Here, $\hat{p}(z_d=k)$ is the estimated value of $\mathbb{E}_{\phi,\psi|W} \big( p(z_d=k | W_d, \phi, \psi) \big)$, computed on the MCMC output by normalising 
\begin{equation} \label{eq:z_d_posterior}
    p(z_d=k | W_d, \phi, \psi) \propto \tilde{\phi}^{\gamma_d,\tau_d}_{k} \prod_{w \in W_d} \tilde{\psi}^{k,\tau_d}_{w}
\end{equation}
over $k \in \{1,\dots,K\}$. Recall that we have $K'$ true senses, whereas we run the models using $K$ senses (with $K \geq K'$), so modelled senses may be grouped together to map them onto the true senses. This is discussed further in Section~\ref{sec:model_selection}. 

\begin{table}[!t]
\caption{Brier scores for test data using different models with HMC sampling. Best scores are in blue, and scores from the models selected using the criteria set out in Section~\ref{sec:model_selection} (so independently of the Brier scores) are boxed. 95\% confidence intervals for the reported scores are typically within $\text{BS} \pm 0.005$.} \label{tab:brier_scores_summary}
\centering
\begin{threeparttable}
\begin{tabular}{l S[table-format=1.3] S[table-format=1.3] S[table-format=1.3] S[table-format=1.3]}
\toprule
                    & bank  & kosmos & mus & harmonia \\
\midrule
Uniform predictions & 0.500 & 0.667 & 0.667 & 0.667 \\
GASC/SCAN           & 0.172 &     * &     * &     * \\
DiSC                & 0.150 & 0.371 & 0.203 & 0.639 \\
EDiSC ($M = 50$)    & 0.139 & 0.349 & 0.135 &     * \\
EDiSC ($M = 100$)   & 0.139 & 0.329 & \boxed{\textcolor{blue}{0.093}} & * \\
EDiSC ($M = 200$)   & \boxed{\textcolor{blue}{0.133}} & \boxed{0.332} & 0.099 & * \\
EDiSC ($M = 300$)   & 0.143 & \textcolor{blue}{0.326} & 0.101 & \boxed{\textcolor{blue}{0.584}} \\
\bottomrule
\end{tabular}
\small{* Not converged: MCMC runs from different starting configurations lead to different equilibrium distributions}
\end{threeparttable}
\end{table}

The results for each model and dataset, obtained using converged HMC runs, are summarised in Table~\ref{tab:brier_scores_summary}. Under uniform predictions, if we set $\hat{p}(z_d=k) = \frac{1}{K'}$ for all $d,k$, we get $\text{BS} = \left(1-\frac{1}{K'}\right)^2 + (K'-1)\left(\frac{1}{K'}\right)^2 = 0.5$ in the case of $K'=2$ for ``bank'' or $0.667$ in the case of $K'=3$ for the other datasets; so models must produce scores lower than these in order to be useful. In all our test cases, EDiSC with an appropriate dimension $M$ offers a clear improvement over DiSC. Recall that DiSC was already an improvement over GASC/SCAN, so we treat DiSC as the only baseline and do not compare against GASC/SCAN hereinafter. 

The extent of improvement provided by EDiSC over DiSC varies depending on the complexity of the target dataset. For the simpler ``bank'' test case, with a large number of snippets relative to the vocabulary size, the two senses are quite distinct and well-informed by the snippet data, so there is limited scope for improvement. For the more challenging Greek test cases, there are underlying semantic relationships in the wider corpus that cannot be captured through the snippet data alone, but are learnt via the context-word embeddings, so the improvement is more pronounced. The improvement is greatest for ``mus'', where the small data size limits DiSC performance, but the inclusion of embeddings in EDiSC counteracts by providing added structure via the learnt context-word relationships latent in the corpus.


Note that our goal is to infer $\tilde{\phi}$ and $\tilde{\psi}$ given the snippets, rather than WSD. We obtain the probabilistic predictions \eqref{eq:z_d_posterior} as a free byproduct from the converged posteriors, which \textit{could} be used for WSD, but we are not in essence trying to tag instances of our target words with their correct sense. Typically for WSD (and related NLP tasks) with a benchmark dataset, the precision, recall and their harmonic mean (F1 score) are used for assessment. These are appropriate when making 0/1 predictions on sense labels. However, in the case of probabilistic predictions such as ours, a scoring rule is more appropriate. Whilst any proper scoring rule could be used for this purpose, some common alternatives being logarithmic or spherical score, we choose the Brier score for its attractive properties: it is equivalent to the widely used and well-understood mean squared error; the contribution from any one prediction is bounded (in contrast to logarithmic score, which can be unstable); and it has symmetrical penalties for over-confidence in the wrong prediction and under-confidence in the right prediction (in contrast to both logarithmic and spherical scores).

\subsection{Model selection} \label{sec:model_selection}

Two important modelling choices to make are the number of model senses $K$ and the embedding dimension $M$. We first consider the choice of $K$.

The models discussed in this paper are useful tools for exploratory analysis of unlabelled snippets and, as such, their success is linked to whether the model output is meaningful to a user. Setting $K$ is like choosing a resolution for how fine we want sense differences to be resolved. This suggests setting $K$ in a semi-supervised mode, where we learn the model parameters unsupervised for a few values of $K$, and the user assigns meaningful labels to the posterior sense distributions based on the model output. A low value of $K$ that is meaningful to the user would help with interpretability, whereas a higher value may fit the data better, and the user can select $K$ based on this trade-off. We demonstrate what this looks like in practice using the ``bank'' example.

\begin{table}[!t]
\caption{``Bank'' top 10 context words for each model sense under EDiSC} \label{tab:bank_top_words}
\centering
\begin{adjustbox}{max width=\textwidth}
\begin{tabular}{c l l l l l l l l l l}
\toprule
Sense & \multicolumn{10}{l}{Top 10 context words for EDiSC $M=200$ with $K=2$ senses} \\
\midrule
1 & river & stream & water & stand & tree & leave & creek & day & land & reach \\
2 & bank & national & note & money & deposit & reserve & credit & saving & loan & federal \\
\toprule
Sense & \multicolumn{10}{l}{Top 10 context words for EDiSC $M=200$ with $K=3$ senses} \\
\midrule
1 & river & stream & water & stand & tree & leave & creek & bank & reach & day \\
2 & note & bank & money & deposit & reserve & credit & issue & federal & account & loan \\
3 & national & bank & saving & company & president & city & loan & banking & trust & institution \\
\bottomrule
\end{tabular}
\end{adjustbox}
\end{table}

A natural way to examine the posterior is to look at the context words $v$ with the highest probabilities $\frac{1}{T} \sum_{t=1}^T \tilde{\psi}^{k,t}_v$ under each model sense $k$, marginally over time. Table~\ref{tab:bank_top_words} shows the top 10 most probable context words for ``bank'' if we run EDiSC with $K=2$ and $K=3$ model senses, using embedding dimension $M=200$ in both cases. With $K=2$, the model senses are readily recognisable as riverbank and financial institution respectively. With $K=3$, sense 1 is recognisable as riverbank, whereas senses 2 and 3 are both recognisable as financial institution. In the latter case, whilst senses 2 and 3 have different distributions, there is some overlap in the most probable words, which is undesirable since we would like the model senses to be as distinct as possible to help with interpretability. We therefore choose $K=2$ in this case, which is the smallest value giving meaningful model output. 

Incidentally, $K=3$ fits the data better on merging the split financial institution senses of bank, both in terms of predictive accuracy and true-model recovery. However, the true labels are not generally available, and therefore cannot be used for model selection. 
Ultimately, choosing $K$ is up to the user's judgement.

Another important consideration (for both $K$ and $M$) is MCMC convergence. As discussed in Section~\ref{sec:inference}, the posterior \eqref{eq:EDiSC_posterior} is challenging to sample due to metastability and multimodality. Choosing and carefully tuning a good sampler may help overcome the metastability, but the posterior may still be multimodal. If we were to condition on the true sense labels, the resulting posterior is unimodal (as it has strongly informative data). We would like to find a model for the unlabelled data that gives a posterior resembling that for the labelled data. We therefore favour models that are more concentrated and unimodal. This can be explored using MCMC. Some configurations of $K$ and $M$ tend to give multimodal posteriors for some datasets. This may indicate model misspecification. Conversely, a model with a unimodal posterior, in which the model senses are interpretable, is indicative of a well-specified model in our setting.

\begin{table}[!t]
\caption{``Kosmos'' two different modes under EDiSC with $M=100$ and $K=3$. Senses 1, 2 and 3 are interpretable as decoration, order and world respectively in the `correct' mode. The three true senses are not distinguishable in the `incorrect' mode.}
\label{tab:kosmos_modes}
\centering
\begin{adjustbox}{max width=\textwidth}
\begin{tabular}{c p{0.095\linewidth} p{0.095\linewidth} p{0.095\linewidth} p{0.095\linewidth} p{0.095\linewidth} p{0.095\linewidth} p{0.095\linewidth} p{0.095\linewidth} p{0.095\linewidth} p{0.095\linewidth} }
\toprule
Sense & \multicolumn{10}{l}{Top 10 context words for each model sense in the `correct' mode} \\
\midrule
1 & \textgreek{ἔχω} & \textgreek{πολύς} & \textgreek{πᾶς} & \textgreek{γυνή} & \textgreek{καλός} & \textgreek{μέγας} & \textgreek{χρύσεος} & \textgreek{φέρω} & \textgreek{γίγνομαι} & \textgreek{κοσμέω} \\
& \footnotesize (to have) & \footnotesize (many, much) & \footnotesize (all) & \footnotesize (woman) & \footnotesize (beautiful) & \footnotesize (large) & \footnotesize (golden) & \footnotesize (to carry) & \footnotesize (become) & \footnotesize (adorn or arrange) \\
2 & \textgreek{πολιτεία} & \textgreek{πᾶς} & \textgreek{τάξις} & \textgreek{γίγνομαι} & \textgreek{ἔρχομαι} & \textgreek{καθίστημι} & \textgreek{πόλις} & \textgreek{πολύς} & \textgreek{πρότερος} & \textgreek{τρέπω} \\
& \footnotesize (citizenship) & \footnotesize (all) & \footnotesize (arrange-ment) & \footnotesize (become) & \footnotesize (to go) & \footnotesize (to set in order) & \footnotesize (city) & \footnotesize (many, much & \footnotesize (before) & \footnotesize (to rotate) \\
3 & \textgreek{πᾶς} & \textgreek{οὐρανός} & \textgreek{θεός} & \textgreek{γῆ} & \textgreek{γίγνομαι} & \textgreek{κόσμος} & \textgreek{λέγω} & \textgreek{ἔχω} & \textgreek{ὅλος} & \textgreek{Ζεύς} \\
& \footnotesize (all) & \footnotesize (sky) & \footnotesize (divine) & \footnotesize (earth) & \footnotesize (become) & \footnotesize (kosmos) & \footnotesize (to say) & \footnotesize (to have) & \footnotesize (entire) & \footnotesize (Zeus) \\
\toprule
Sense & \multicolumn{10}{l}{Top 10 context words for each model sense in the `incorrect' mode} \\
\midrule
1 & \textgreek{πᾶς} & \textgreek{ἔχω} & \textgreek{πολύς} & \textgreek{γίγνομαι} & \textgreek{γυνή} & \textgreek{καλός} & \textgreek{φέρω} & \textgreek{μέγας} & \textgreek{πόλις} & \textgreek{χρύσεος} \\
& \footnotesize (all) & \footnotesize (to have) & \footnotesize (many, much) & \footnotesize (become) & \footnotesize (woman) & \footnotesize (beautiful) & \footnotesize (to carry) & \footnotesize (large) & \footnotesize (city) & \footnotesize (golden) \\
2 & \textgreek{θεός} & \textgreek{πατήρ} & \textgreek{κύριος} & \textgreek{κόσμος} & \textgreek{οὐρανός} & \textgreek{πᾶς} & \textgreek{υἱός} & \textgreek{Ἰησοῦς} & \textgreek{εἶπον} & \textgreek{αἰών}   \\
& \footnotesize (divine) & \footnotesize (father) & \footnotesize (ruling, lord) & \footnotesize (kosmos) & \footnotesize (sky) & \footnotesize (all) & \footnotesize (son) & \footnotesize (Jesus) & \footnotesize (to speak) & \footnotesize (lifetime, epoch) \\
3 & \textgreek{πᾶς} & \textgreek{οὐρανός} & \textgreek{γῆ} & \textgreek{ἔχω} & \textgreek{γίγνομαι} & \textgreek{λέγω} & \textgreek{ὅλος} & \textgreek{κόσμος} & \textgreek{φημί} & \textgreek{φύσις}  \\
& \footnotesize (all) & \footnotesize (sky) & \footnotesize (earth) & \footnotesize (to have) & \footnotesize (become) & \footnotesize (to say) & \footnotesize (entire) & \footnotesize (kosmos) & \footnotesize (to speak) & \footnotesize (origin) \\
\bottomrule
\end{tabular}
\end{adjustbox}
\end{table}

\begin{table}[!t]
\caption{``Kosmos'' top 10 context words for each model sense under EDiSC. Sense 1 corresponds to decoration, sense 2 to order, and senses 3 and 4 to world.} 
\label{tab:kosmos_top_words}
\centering
\begin{adjustbox}{max width=\textwidth}
\begin{tabular}{c p{0.095\linewidth} p{0.095\linewidth} p{0.095\linewidth} p{0.095\linewidth} p{0.095\linewidth} p{0.095\linewidth} p{0.095\linewidth} p{0.095\linewidth} p{0.095\linewidth} p{0.095\linewidth} }
\toprule
Sense & \multicolumn{10}{l}{Top 10 context words for EDiSC $M=100$ with $K=4$ senses} \\
\midrule
1 & \textgreek{ἔχω} & \textgreek{πᾶς} & \textgreek{πολύς} & \textgreek{γυνή} & \textgreek{καλός} & \textgreek{μέγας} & \textgreek{φέρω} & \textgreek{χρύσεος} & \textgreek{κοσμέω} & \textgreek{γίγνομαι} \\
& \footnotesize (to have) & \footnotesize (all) & \footnotesize (many, much) & \footnotesize (woman) & \footnotesize (beautiful) & \footnotesize (large) & \footnotesize (to carry) & \footnotesize (golden) & \footnotesize (adorn or arrange) & \footnotesize (become) \\
2 & \textgreek{πᾶς} & \textgreek{πολιτεία} & \textgreek{τάξις} & \textgreek{γίγνομαι} & \textgreek{ἔρχομαι} & \textgreek{καθίστημι} & \textgreek{πόλις} & \textgreek{τρέπω} & \textgreek{πολύς} & \textgreek{πολέμιος} \\
& \footnotesize (all) & \footnotesize (citizenship) & \footnotesize (arrange-ment) & \footnotesize (become) & \footnotesize (to go) & \footnotesize (to set in order) & \footnotesize (city) & \footnotesize (to rotate) & \footnotesize (many, much) & \footnotesize (belonging to war) \\
3 & \textgreek{πᾶς} & \textgreek{γῆ} & \textgreek{οὐρανός} & \textgreek{ἔχω} & \textgreek{γίγνομαι} & \textgreek{λέγω} & \textgreek{ὅλος} & \textgreek{κόσμος} & \textgreek{φύσις} & \textgreek{ἕκαστος} \\
& \footnotesize (all) & \footnotesize (earth) & \footnotesize (sky) & \footnotesize (to have) & \footnotesize (become) & \footnotesize (to say) & \footnotesize (entire) & \footnotesize (kosmos) & \footnotesize (origin) & \footnotesize (each, every) \\
4 & \textgreek{θεός} & \textgreek{πατήρ} & \textgreek{οὐρανός} & \textgreek{κόσμος} & \textgreek{κύριος} & \textgreek{πᾶς} & \textgreek{εἶπον} & \textgreek{λέγω} & \textgreek{ἔρχομαι} & \textgreek{υἱός} \\
& \footnotesize (divine) & \footnotesize (father) & \footnotesize (sky) & \footnotesize (kosmos) & \footnotesize (ruling, lord) & \footnotesize (all) & \footnotesize (to speak) & \footnotesize (to say) & \footnotesize (to go) & \footnotesize (son) \\
\bottomrule
\end{tabular}
\end{adjustbox}
\end{table}

As an example, running EDiSC with $M=100$ and $K=3$ on the ``kosmos'' data, the samplers settle in one of two distinct modes as shown in Table~\ref{tab:kosmos_modes}. The likelihood is the same in both cases. In the `correct' mode, the senses are recognisable due to representative words like \textgreek{γυνή} (woman) and \textgreek{χρύσεος} (golden) for the ``decoration'' sense, \textgreek{πολιτεία} (citizenship) and \textgreek{τάξις} (arrangement) for the ``order'' sense, and \textgreek{γῆ} (earth) and \textgreek{οὐρανός} (sky) for the ``world'' sense. However, in the `incorrect' mode, the ``order'' and ``world'' senses do not appear separated, with some words like \textgreek{οὐρανός} (sky) appearing with high probability under both model senses. The Brier scores reflect this: $\text{BS} = 0.322$ in the first case and $\text{BS} = 0.620$ in the second case (the best score obtained through all possible mappings of model senses to true senses). Note that all Greek translations have been obtained from Wiktionary, and we have only included a few representative meanings to help the reader follow.

In general, a model with a multimodal posterior should be avoided since we cannot be sure which (if any) is the `true' mode. We find that we get unimodal posteriors with $K=4$ for the ``kosmos'' data, $K=3$ for the ``mus'' data, and $K=4$ for the ``harmonia'' data. These are also the lowest $K$ values for which the model senses are interpretable, and so we set $K$ accordingly. With ``kosmos'', two of the model senses map to the ``world'' sense, and an example for $M=100$ is shown in Table~\ref{tab:kosmos_top_words}. Similarly, with ``harmonia'', two of the model senses map to the ``abstract'' sense in the converged runs.

In choosing $M$, the first consideration should be MCMC convergence as discussed above. This is because if the model fails to converge for certain $M$ values, it may be that the embeddings learnt at that dimension do not capture the context-word semantics adequately for our purpose, and so the model is misspecified. After excluding the non-converging settings, model-selection tools may be used. However, the choice of $M$ is ultimately up to the user's judgement, so other factors such as marginal gains or computational costs may be considered. Note that $M$ cannot be decided on the fly: the embeddings must be learnt separately for fixed values of $M$. Therefore, practically, the choice has to be made out of a handful of predetermined values, which in our case are 50, 100, 200 and 300. 

\begin{figure}[!t]
\centering
\makebox[\textwidth][c]{\includegraphics[width = 1.1\textwidth, keepaspectratio]{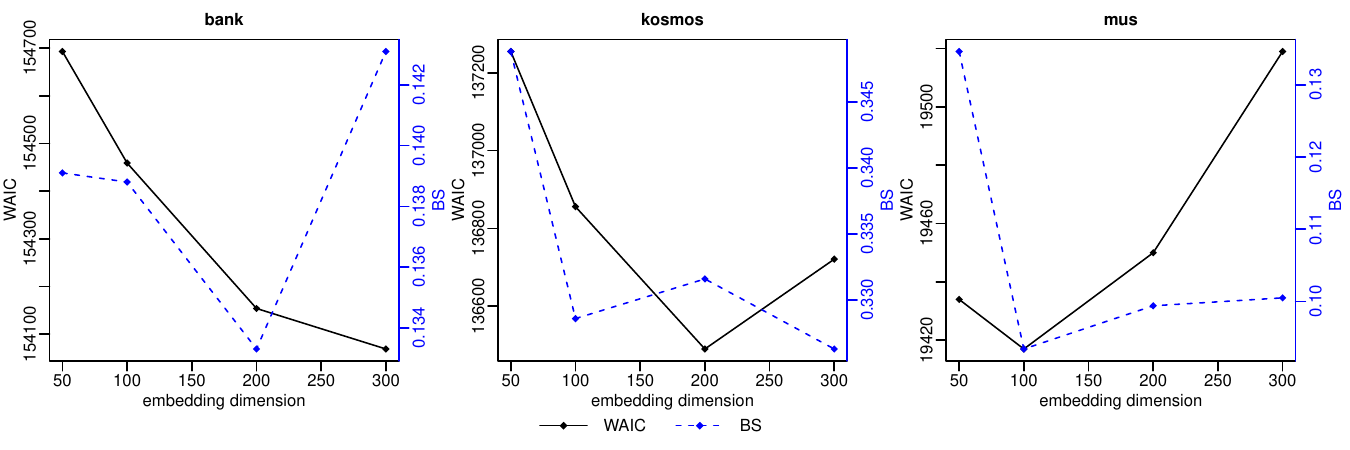}}
\vspace{-20pt}
\caption{WAIC and Brier scores for different choices of embedding dimension $M$ for the ``bank'', ``kosmos'' and ``mus'' data}
\label{fig:WAIC}
\end{figure}

We use the widely applicable information criterion (WAIC) \citep{watanabe2010asymptotic, vehtari2017practical} to guide our choice as it is a computationally convenient model selection tool for Bayesian inference. It has several slightly different formulations, but the one used in the R packages \texttt{LaplacesDemon} \citep{R_LaplacesDemon} and \texttt{loo} \citep{R_loo} is
\begin{align} \label{eq:WAIC}
    \text{WAIC} &= -2 (\widehat{\text{LPD}} - \hat{p}_\text{WAIC})  \nonumber \\
    &= -2 \sum_{d=1}^D \Big( \log \mathbb{E}_{\phi,\psi|W} [p(W_d | \phi, \psi)]  - \mathbb{V}_{\phi,\psi|W} [\log p(W_d | \phi, \psi)] \Big) \text{,}
\end{align}
where $\widehat{\text{LPD}}$ estimates the log pointwise predictive density (LPD), $\hat{p}_\text{WAIC}$ estimates the effective number of parameters, and $(\widehat{\text{LPD}} - \hat{p}_\text{WAIC})$ estimates the expected log pointwise predictive density (ELPD). The WAIC is a predictive loss like the Akaike information criterion (AIC), and asymptotically equivalent to selecting the model that maximises the posterior predictive probability density for held-out data in a leave-one-out cross-validation (LOOCV) setup.

Figure~\ref{fig:WAIC} shows how the WAIC varies with the choice of $M$, and also shows the Brier scores for these choices, for the ``bank'', ``kosmos'' and ``mus'' data. There seems to be a loose correlation between the WAIC and BS, which suggests that the WAIC is a sensible tool to use when validation data (with true sense labels) is not available. For the ``kosmos'' and ``mus'' data, the WAIC is minimised at $M=200$ and $M=100$ respectively, so we go with these choices. For the ``bank'' data, the WAIC does not seem to have a local minimum within our range. On the other hand, the marginal gains between $M=200 \text{ and } 300$ are relatively low, whereas the computational cost is much higher. We therefore select $M=200$ for ``bank'', which strikes a good balance. For ``harmonia'', we only get MCMC convergence for $M=300$, so that is the only choice.

One may ask why we use the WAIC to guide our choice of $M$ but not of $K$. This is because, as opposed to $M$, the WAIC always favours a higher $K$ value in our experiments. This, in turn, is because of how these two parameters interact with the likelihood \eqref{eq:likelihood}: $K$ directly changes the number of variables used in the likelihood calculation via the dimensions of $\tilde{\phi}$ and $\tilde{\psi}$, whereas $M$ only indirectly affects the likelihood via the relation $\tilde{\psi}^{k,t} = \softmax \left( \rho (\chi^k + \theta^t) + \varsigma \right)$ without changing the dimensions of $\tilde{\psi}$ itself. \citet{vehtari2017practical} state that $\hat{p}_\text{WAIC}$ in \eqref{eq:WAIC} can be severely understated in case of a weak prior, and is unreliable if any of the terms $\mathbb{V}_{\phi,\psi|W} [\log p(W_d | \phi, \psi)]$ exceed 0.4, which is frequently the case in our experiments. This is not a problem when choosing $M$, as the effective dimension, estimated by the variance term, does not change much from one $M$-value to another; so the order of the models is decided mainly by goodness of fit $\widehat{\text{LPD}}$, and the unreliable dimension estimate $\hat{p}_\text{WAIC}$ has little impact. However, model selection for $K$ is a trade-off between model fit (LPD) and parsimony (variance); so the poor estimate $\hat{p}_\text{WAIC}$ of the effective dimension is an obstacle. In any case, the WAIC should be used as a guide rather than a definitive rule.

\subsection{Sense-prevalence estimation} \label{sec:sense_prevalence_estimation}

Predictive accuracy and true-model recovery are the two classical goals of statistical inference. However, good performance on one front does not necessarily correlate with good performance on the other. We examined predictive performance using the Brier score. We now assess our fitted models against the `true' models.

We have the posterior sense-prevalence distributions $\tilde{\phi}|W$ given the \textit{unlabelled} data $W$. We use the 95\% highest posterior density (HPD) intervals as concise visual summaries of the support for the posterior and the uncertainty in the estimates. Following \citet[Section~7]{DiSC_https://doi.org/10.1111/rssc.12591}, we would like to compare these HPDs to the unknown true sense prevalence, $\tilde{\Phi}$ say. We do not have $\tilde{\Phi}$: we only have the true sense labels $o_{1:D}$. However, given $o_{1:D}$, estimating $\tilde{\Phi}$ with uncertainty quantification is an easy and classical task: we simply smooth the empirical sense-prevalence probabilities using a time series model for their evolution. This gives us well-calibrated independent estimates $\tilde{\phi}|(z=o)$ given the \textit{labelled} data $o$, which we use as proxies for $\tilde{\Phi}$. The posteriors $\tilde{\phi}|(z=o)$ would be concentrated on the empirical sense-prevalence probabilities where there are many observations, and would apply some shrinkage and smoothing where snippets are infrequent. 
If the credible sets from the unlabelled analysis are close to the credible sets from the labelled analysis using the same AR(1) models, this would indicate success: conditioning on the true labels gives the best results achievable with these models. 

\begin{figure}[!t]
\centering
\includegraphics[width = 0.9\textwidth, keepaspectratio]{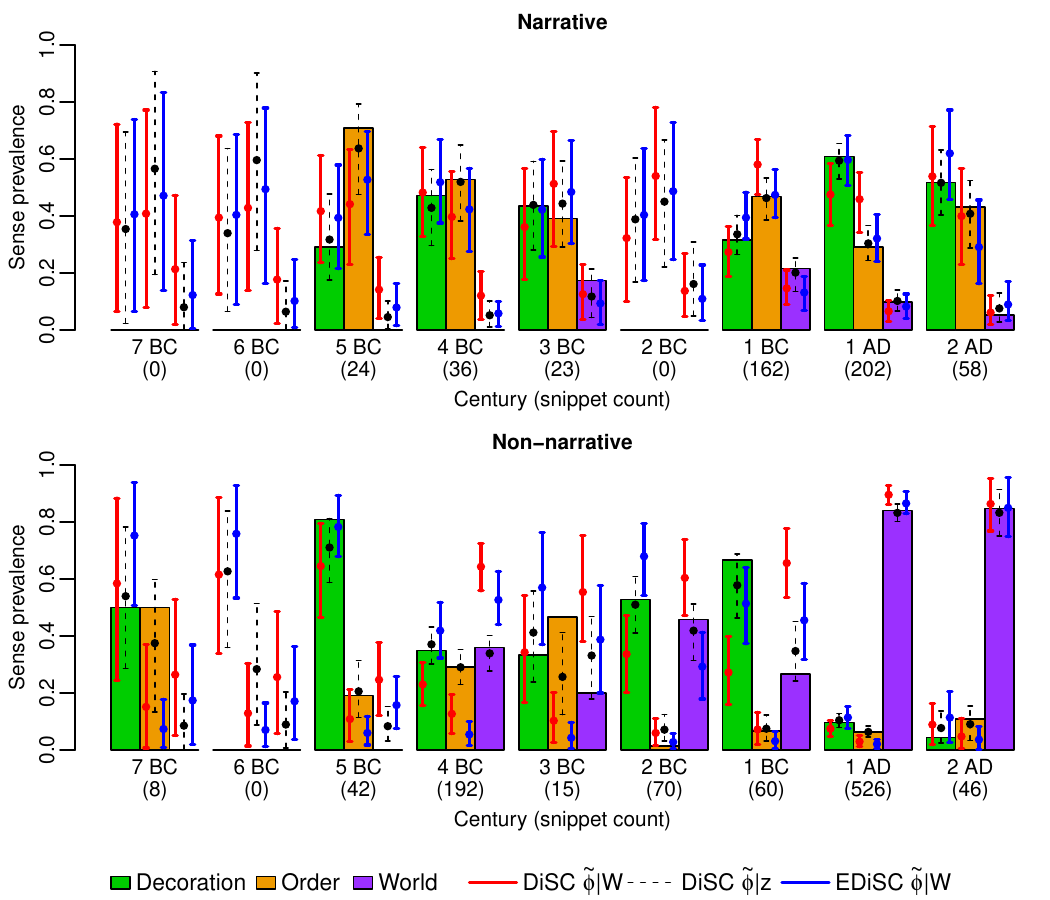}
\caption{``Kosmos'' expert-annotated empirical sense prevalence (coloured bars with height $N_{k,g,t}^{o} / \sum_{l=1'}^{K'} N_{l,g,t}^{o}$ for each $k,g,t$), and 95\% HPD intervals (error bars) and posterior means (circles) from the model output. Snippet counts $N_{\cdot,g,t}^{o}$ are given in brackets below the axes. Note that the labelled posteriors $\tilde{\phi}|z$ from DiSC and EDiSC are identical.}
\label{fig:kosmos_phi_error_bars}
\end{figure}

We show this comparison for both DiSC and EDiSC posteriors on the ``kosmos'' data in Figure~\ref{fig:kosmos_phi_error_bars}, with the models selected as discussed in Section~\ref{sec:model_selection}. We see that both unlabelled posteriors $\tilde{\phi}|W$ (solid error bars) are in surprisingly good agreement with the labelled posterior $\tilde{\phi}|(z=o)$ (dashed error bars). However, EDiSC has generally better range and location: the blue EDiSC error bars generally have higher overlap with the dashed bars, and the circles (posterior means) are closer. The unlabelled analysis is equivalent to many labelled analyses averaged over uncertainty in reconstructed labels. This uncertainty is significant, so it is remarkable how close the unlabelled analysis comes to the labelled analysis. It is also interesting that particular features of the ground truth are reflected well in the posteriors, such as the emergence of the ``world'' sense of ``kosmos'' in 4th century BC. 

\begin{table}[!t]
\caption{\label{tab:BF_kosmos}``Kosmos'' Bayes factors $\text{BF}_{01}$ on a $\log_{10}$ scale for nested `true' model $H_0: \tilde{\phi}^{g,t} \in \mathcal{S}^{g,t}$ over each of $H_1:$ DiSC and $H_1:$ EDiSC. \textcolor{red}{Red} indicates incorrect rejection of $H_0$.}
\centering
\begin{tabular}{ l l rrrrrrrrr}
    \toprule
    Model & Genre & 7 BC & 6 BC & 5 BC & 4 BC & 3 BC & 2 BC & 1 BC & 1 AD & 2 AD \\
    \midrule
    DiSC  & narrative & $0.28$ & $0.41$ & $0.47$ & $0.79$ & $0.86$ & $0.62$ & $0.77$ & $0.18$ & $1.20$ \\
    EDiSC & narrative & $0.43$ & $0.63$ & $1.06$ & $1.20$ & $0.92$ & $0.62$ & $1.09$ & $1.63$ & $0.97$ \\
    \midrule
    DiSC  & non-narr & $-0.05$ & $0.08$ & $-0.65$ & $\textcolor{red}{-\infty}$ & $-0.26$ & $0.41$ & $\textcolor{red}{-1.54}$ & $0.44$ & $1.10$ \\
    EDiSC & non-narr & $-0.35$ & $0.01$ & $-0.36$ & $\textcolor{red}{-\infty}$ & $-0.99$ & $0.37$ & $0.77$ & $-0.07$ & $0.92$ \\
    \bottomrule
\end{tabular}
\end{table}

The figure shows the empirical sense-prevalence estimates $N_{k,g,t}^{o} / N_{\cdot,g,t}^{o}$ as coloured bars, where $N_{k,g,t}^{o}$ is the count of snippets in genre $g$ at time $t$ with sense $k$ under the true sense labels $o$, and $N_{\cdot,g,t}^{o} = \sum_{l=1'}^{K'} N_{l,g,t}^{o}$ is the total snippet count in each $g,t$ (shown in brackets below the axes). The empirical estimates may be of interest, and are shown on the same figure as they are defined on the same space and scale. However, the empirical estimates should not be used to assess true-$\tilde{\Phi}$ recovery if the datasets are small and sparse, as is the case with our ancient Greek data, since they are not smoothed.

To make the comparison more concrete, we quantify both models' performance using Bayes factors to measure agreement between the unlabelled and labelled posteriors. We treat the credible sets from the labelled analysis as the truth and ask, does the unlabelled analysis reject the truth? Let $\mathcal{S}^{g,t}$ be the 95\% HPD region for genre $g$ and time $t$ under the labelled posterior $\tilde{\phi}^{g,t}|(z=o)$. For each $g$ and $t$, we compute the Bayes factor $\text{BF}_{01} = \frac{p(W|H_0)}{p(W|H_1)}$ for the nested `true' model $H_0: \tilde{\phi}^{g,t} \in \mathcal{S}^{g,t}$ over each of $H_1:$ DiSC and $H_1:$ EDiSC. By the Savage-Dickey density ratio, we have 
\begin{equation*}
    \text{BF}_{01} = \frac{\pi(\tilde{\phi}^{g,t} \in \mathcal{S}^{g,t} | W, H_1)}{\pi(\tilde{\phi}^{g,t} \in \mathcal{S}^{g,t} | H_1)} \text{,}
\end{equation*}
where the posterior probabilities in the numerator can be computed using our MCMC samples from the unlabelled $\tilde{\phi}^{g,t}|W$ posteriors, and the prior probabilities in the denominator can be computed using softmax-transformed Monte Carlo simulations from the prior $\phi^{g,t}$. We use the same prior simulations for DiSC and EDiSC so that the model comparison is indifferent to simulation error in the denominator. The results are reported in Table~\ref{tab:BF_kosmos} on a $\log_{10}$ scale. Positive values indicate evidence in favour of $H_0$, with higher $\text{BF}_{01}$ corresponding to greater overlap between $\tilde{\phi}^{g,t}|W$ and $\tilde{\phi}^{g,t}|(z=o)$, and vice versa for negative values. Using the scale in \citet{doi:10.1080/01621459.1995.10476572}, $\log_{10}(\text{BF}_{01}) < -1$ indicates strong evidence against $H_0$; so we reject $H_0$ in favour of $H_1$ at this threshold. These rejections are highlighted in \textcolor{red}{red} in the table. We find that EDiSC gives a higher Bayes factor than DiSC in 61\% of cases, and incorrectly rejects $H_0$ only once versus twice for DiSC. Further, when EDiSC performs better, the improvement can be substantial, for example 1~AD in the narrative genre. The converse is not true: when DiSC performs better, the difference is only slight. 

Equivalent figures and tables for the ``mus'', ``harmonia'' and ``bank'' data are given in Appendix~\ref{sec:further_results}. These similarly show EDiSC outperforming DiSC on true-$\tilde{\Phi}$ recovery in 56, 83 and 90 percent of cases respectively, with fewer incorrect rejections of $H_0$, less bias, and more accurate and precise credible sets.

\subsection{Sampling efficiency and scalability} \label{sec:sampling}

We assess the relative efficiency of sampling the DiSC and EDiSC posteriors using MCMC methods. The form of the model $\psi^{k,t} = \rho (\chi^k + \theta^t) + \varsigma$ in EDiSC, compared to $\psi^{k,t} = \chi^k + \theta^t$ in DiSC, means that additional matrix multiplication is required to calculate the likelihood for EDiSC, which tends to be computationally slow. However, using the effective sample size (ESS) per unit time as a metric for comparing the sampling efficiency, we find that, in fact, the same MCMC applied to EDiSC results in more efficient samples than DiSC. This is because of the lower dimensional model parameters $\chi, \theta$ for EDiSC.

\begin{table}[!t]
\caption{\label{tab:ess_bank}Median (interquartile range) ESS per hour of CPU time from MALA sampling}
\centering
\begin{tabular}{ l r c r c }
    \toprule
    Model & \multicolumn{2}{l}{ESS for $\tilde{\phi}$} & \multicolumn{2}{l}{ESS for $\tilde{\psi}$} \\
    \midrule
    DiSC            &   375 &     (216 – 531) & 391 & (213 – 586) \\
    EDiSC ($M=50$)  & 1,916 & (1,569 – 2,017) & 391 & (294 – 515) \\
    EDiSC ($M=100$) & 2,192 & (1,844 – 2,674) & 344 & (262 – 459) \\
    EDiSC ($M=200$) & 2,237 & (1,810 – 2,424) & 303 & (215 – 396) \\
    EDiSC ($M=300$) & 1,633 & (1,245 – 2,299) & 282 & (217 – 392) \\
    \bottomrule
\end{tabular}
\end{table}

Table~\ref{tab:ess_bank} shows the ESS per hour of CPU time from applying MALA to target the $\tilde{\phi}$ and $\tilde{\psi}$ posteriors under DiSC and EDiSC on the ``bank'' data. We report the medians and interquartile ranges over all the $KGT$ parameters for $\tilde{\phi}$, and over the top 20 most probable words (i.e. over $20KT$ parameters) out of the $VKT$ parameters for $\tilde{\psi}$. We compare implementations not algorithms. However, the comparison is fair as the MALA Monte Carlo is common to both, and the evaluation allows little scope for differential optimisation of implementations. (For example, MALA requires no optimisation of the number of leapfrog steps, as opposed to HMC or NUTS.) All runs were done sequentially on the same PC. We see that whilst the ESS for $\tilde{\psi}$ is of the same order for all models, the ESS for $\tilde{\phi}$ is many times better under EDiSC than under DiSC. 

\begin{figure}[t]
\centering
\includegraphics[width = 1\textwidth, keepaspectratio]{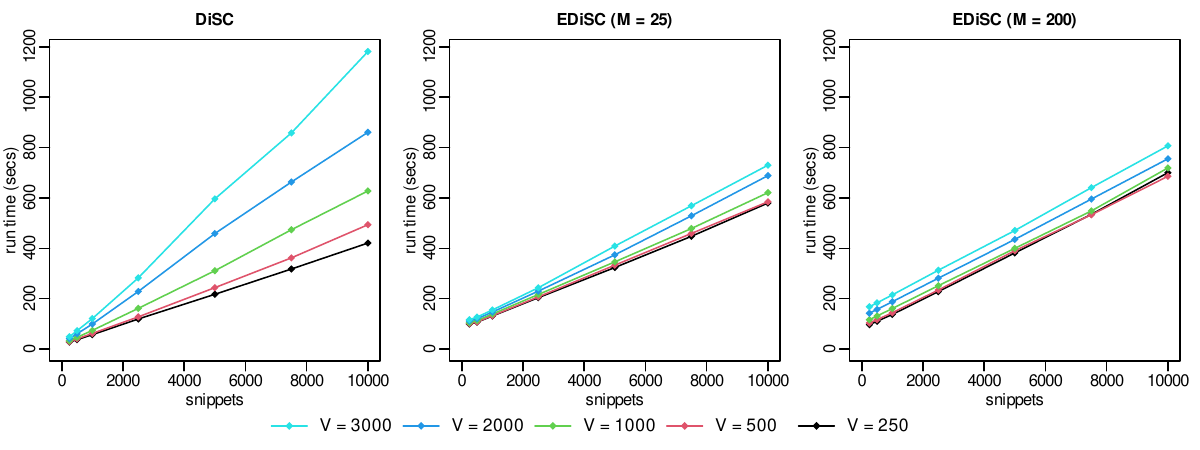}
\vspace{-20pt}
\caption{Mean run times in CPU seconds for 500 MCMC iterations on synthetic data using different models, vocabulary sizes ($V$) and number of snippets ($D)$}
\label{fig:run_times}
\end{figure}

We also analyse how the run times vary with increasing data sizes using synthetic data. Figure~\ref{fig:run_times} summarises the results, where each data point is the mean run-time over three independent runs of 500 MCMC iterations. This comparison favours DiSC, since we use the same number of MCMC iterations for DiSC and EDiSC despite EDiSC being more efficient. We show the results from two different choices for the embedding dimension for EDiSC: $M=25$ and $M=200$. Typically, the vocabulary size $V$ would be expected to grow with the data size $D$ (as more snippets tend to bring a larger vocabulary), whereas the embedding dimension would not usually be increased much beyond 200.

The run times increase linearly with both vocabulary size $V$ and number of snippets $D$ in all cases. However, for $V>500$ (which is typical) the rate of increase with $D$ is much higher for DiSC than for EDiSC. The interaction effect between $V$ and $D$ on the run time is significant in all cases. However, the interaction effect is much stronger for DiSC than for EDiSC. Moreover, as the embedding dimension $M$ is increased, the interaction effect for EDiSC grows even weaker. As a result, with increasing $V$ and $D$, run times for DiSC increase much faster than those for EDiSC. The plots show the advantage of using EDiSC over DiSC even in the case of modest data sizes in our synthetic data experiments. Therefore, EDiSC is a lot better suited than DiSC to scaling up for larger data sizes. 

This paper focuses on modelling diachronic sense change for our ancient Greek data, where computational issues are a relatively minor concern. However, the models themselves are more widely applicable. Hence, in other situations where efficiency and scalability are more of an issue, these experiments show that there is even more reason to prefer EDiSC over DiSC (or GASC/SCAN for that matter).

\section{Discussion} \label{sec:discussion}

We introduced EDiSC, an embedded version of DiSC, which is a generative model of diachronic sense change that combines word embeddings with DiSC. Experiments on test data show that EDiSC outperforms DiSC, GASC and SCAN, as measured by Brier scores, in terms of accuracy in predicting the unknown sense for the target word in snippets. We estimated the model parameters and quantified the uncertainty in sense-change estimates via HPD intervals, showing that EDiSC outperforms DiSC on recovering the true parameters: estimates obtained using EDiSC on unlabelled data are more closely aligned to those obtained on labelled data than is the case for DiSC. The good agreement between our HPD intervals computed using unlabelled and labelled data supports our view that it would be very hard to do much better than we have done here, at least in a bag-of-words setting. 

We showed that MCMC sampling targeting EDiSC is more efficient than the corresponding sampling for DiSC. Furthermore, EDiSC scales better to large data sizes than DiSC. We considered how fitting these models is challenging due to potential metastability and multimodality in the posteriors, and why variational methods for model fitting fail. We discussed ways of addressing these challenges that work well in our experiments. These include careful model selection with respect to the number of model senses $K$ and the embedding dimension $M$, as well as MCMC considerations (discussed in the Appendix). More broadly, we gave guidelines for how appropriate values for $K$ and $M$ may be set, such that these meet the user's objectives.

An obvious limitation in the EDiSC model is that it uses traditional word embeddings with only one vector representation per word. Our work is a continuation of existing models (SCAN, GASC, DiSC) using the same general framework, which have previously been used to analyse the ancient Greek data that inspired our research. This framework does not readily admit contextualised word embeddings. It would be interesting to investigate how the framework might be expanded or modified to admit multiple vector representations per word. 

Another (necessary) limitation of our work is that we are restricted in what comparisons we can make against other models and methods. As discussed in the literature review, most other methods use some form of pre-training or supervised learning, and in any case have different modelling and inferential goals to ours. However, our models can be generalised and used for wider purposes such as WSD or sense change-point detection if desired; so it would be interesting to investigate how they compare against other methods on shared tasks, or how they might be used in conjunction with other methods from the NLP literature. That is future work for us.

We set out to develop an embedded version of DiSC to model diachronic sense change for our ancient Greek data, drawing parallels from the topic modelling literature, which was an improvement upon the existing model. This objective has been achieved, and prospects for further improvement are good.

\section*{Implementation}
The code and data used to produce the results reported in this paper are available from \url{https://github.com/schyanzafar/EDiSC}.



\section*{Declaration of conflicting interests}
The authors declared no potential conflicts of interest with respect to the research, authorship and/or publication of this article.

\section*{Funding}
This research is funded by the Engineering and Physical Sciences Research Council (EPSRC) under grant EP/S515541/1.

\appendix
\section*{Appendix}

\section{Hyperparameter settings} \label{sec:hyperparameters}


For the AR(1) process hyperparameters $\alpha_\phi, \alpha_\theta$ in EDiSC, a high value admits weak mean reversion without unduly influencing the posteriors. In our model fitting, we therefore experimented with values of 0.9 as in DiSC, and even higher values of 0.99. We found the converged posteriors to be robust to these choices. However, with $\alpha_\theta = 0.99$, convergence problems become more frequent. Moreover, using the WAIC for model selection, values of 0.9 are preferred. Therefore, we continue to use $\alpha_\phi = \alpha_\theta = 0.9$ for EDiSC.

For the $\kappa_\phi$ variance hyperparameter, \citet[Section~4.2]{DiSC_https://doi.org/10.1111/rssc.12591} elicit a prior by defining what we consider to be an extreme (i.e. 3-sigma) event, and using this to set quantiles. For any fixed time $t$, genre $g$ and pair of senses $l,m  \in \{1,\dots,K\}$, a difference in sense prevalence of the order $\tilde{\phi}^{g,t}_l / \tilde{\phi}^{g,t}_m \approx 100$ is considered extreme. Therefore, on the log scale, $\phi^{g,t}_l - \phi^{g,t}_m > \log 100$ is considered a 3-sigma event. Since $\mathbb{V} ( \phi^{g,t}_l - \phi^{g,t}_m ) = \frac{2\kappa_\phi}{1-(\alpha_\phi)^2}$, we express our preference with $3 \left( \frac{2\kappa_\phi}{1-(\alpha_\phi)^2} \right)^\frac{1}{2} = \log 100$, giving $\kappa_\phi = \frac{1-(\alpha_\phi)^2}{2} (\frac{1}{3} \log 100)^2 \approx 0.25$ on rounding. The $\phi$ parameter in EDiSC is identical to that in DiSC, so we continue to use the same value.

For the other variance hyperparameters, still following \citet[Section~4.2]{DiSC_https://doi.org/10.1111/rssc.12591}, for any fixed time $t$, sense $k$ and pair of words $x,y \in \{1,\dots,V\}$, the ratio of context-word probabilities of the order $\tilde{\psi}^{k,t}_x / \tilde{\psi}^{k,t}_y \approx 1\,000$ is considered extreme. Therefore, on the log scale, $\psi^{k,t}_x - \psi^{k,t}_y > \log 1\,000$ is considered a 3-sigma event. 

Now, for EDiSC, we have $\mathbb{V} ( \psi^{k,t}_x - \psi^{k,t}_y ) = \mathbb{V} ( \rho_x^\text{T} \xi^{k,t} - \rho_y^\text{T} \xi^{k,t} + \varsigma_x - \varsigma_y)$, which simplifies to $\mathbb{V} ( \psi^{k,t}_x - \psi^{k,t}_y ) = (\rho_x - \rho_y)^\text{T} \mathbb{V} (\xi^{k,t}) (\rho_x - \rho_y) + \mathbb{V} (\varsigma_x - \varsigma_y)$ since $\xi$ and $\varsigma$ are independent of each other by construction. We have $\mathbb{V} (\varsigma_x - \varsigma_y) = 2 \kappa_\varsigma$ whereas $\mathbb{V} (\xi^{k,t}) = \mathbb{V} (\chi^k + \theta^t)$ is an $M \times M$ diagonal matrix with entries $\left( \kappa_\chi + \frac{\kappa_\theta}{1-(\alpha_\theta)^2} \right)$. Hence, \[\mathbb{V} ( \psi^{k,t}_x - \psi^{k,t}_y ) = (\rho_x - \rho_y)^\text{T} (\rho_x - \rho_y) \left( \kappa_\chi + \frac{\kappa_\theta}{1-(\alpha_\theta)^2} \right) + 2 \kappa_\varsigma.\] We approximate $(\rho_x - \rho_y)^\text{T} (\rho_x - \rho_y)$ with its median $c$ over all pairs $x,y \in \{1,\dots,V\}$, and express our preference with $3 \left( c ( \kappa_\chi + \frac{\kappa_\theta}{1-(\alpha_\theta)^2} ) + 2 \kappa_\varsigma \right)^\frac{1}{2} = \log 1\,000$.

We want the bulk of the variance to be explained by $\xi$, since $\varsigma$ is only a correction parameter that comes into play when the embeddings for words $x,y$ are similar but the words occur at different frequencies in the snippets. Taking the extreme case $\rho_x=\rho_y$, we would not expect the frequency of $x$ to be too different from that of $y$; so we assert that $\psi^{k,t}_x - \psi^{k,t}_y > \log 10$ is a 3-sigma event in this extreme case. Also, $\rho_x=\rho_y$ gives $\mathbb{V} ( \psi^{k,t}_x - \psi^{k,t}_y ) = 2\kappa_\varsigma$, so we express our preference with $3 (2\kappa_\varsigma)^\frac{1}{2} = \log 10$, giving $\kappa_\varsigma = \frac{1}{2} (\frac{1}{3} \log 10)^2 \approx 0.25$ on rounding.

We have $\xi^{k,t} = \chi^k + \theta^t$, so $\mathbb{V} (\xi^{k,t})$ must be apportioned between $\mathbb{V} (\chi^k)$ and $\mathbb{V} (\theta^t)$. Given our preference $\kappa_\chi + \frac{\kappa_\theta}{1-(\alpha_\theta)^2} = \left( (\frac{1}{3}\log 1\,000)^2 - 2\kappa_\varsigma \right) / c$, we set $\kappa_\chi = a_\chi \left( (\frac{1}{3}\log 1\,000)^2 - 2\kappa_\varsigma \right) / c $ and $\kappa_\theta = a_\theta (1-(\alpha_\theta)^2) \left( (\frac{1}{3}\log 1\,000)^2 - 2\kappa_\varsigma \right) / c$, with $a_\chi + a_\theta = 1$. A plausible choice is $a_\chi = a_\theta = 0.5$ as for DiSC, since $\chi$ and $\theta$ are additive effects on the same scale. We experimented with this choice, as well as $a_\chi = 0.75, a_\theta = 0.25$ and $a_\chi = 0.25, a_\theta = 0.75$, and found the posteriors to be robust to these choices. Moreover, the WAIC was relatively constant between these choices, so we go with $a_\chi = a_\theta = 0.5$ for simplicity and consistency with DiSC. However, users may adjust the proportions depending on how they want to resolve the variation over senses and time periods for the test case in question. With these choices, and with $\kappa_\varsigma = 0.25$, we get $\kappa_\chi \approx 2.5/c$ and $\kappa_\theta \approx 0.5/c$ on rounding.

To summarise, for EDiSC, we use $\alpha_\phi = \alpha_\theta = 0.9, \kappa_\phi = 0.25, \kappa_\chi = 2.5/c, \kappa_\theta = 0.5/c, \kappa_\varsigma = 0.25$, where $c$ is the median value of $(\rho_x - \rho_y)^\text{T} (\rho_x - \rho_y)$ over all pairs $x,y \in \{1,\dots,V\}$. For comparison, DiSC uses $\alpha_\phi = \alpha_\theta = 0.9, \kappa_\phi = 0.25, \kappa_\chi = 1.25, \kappa_\theta = 0.25$.

Note that, for GASC and SCAN, the hyperparameters to set are $\kappa_\psi$ and the $a,b$ in $\kappa_\phi \sim \InvGamma(a,b)$. The authors of GASC preferred $\kappa_\psi=0.01,a=1,b=1$ for the Greek test cases, whereas the authors of SCAN preferred $\kappa_\psi=0.1,a=7,b=3$ for their English test cases. In our comparisons, we try both configurations in our implementation of GASC for the Greek test cases, and stick to the SCAN configuration for our English test case.


\section{Further results} \label{sec:further_results}

\begin{figure}[!t]
\centering
\includegraphics[width = 0.9\textwidth, keepaspectratio]{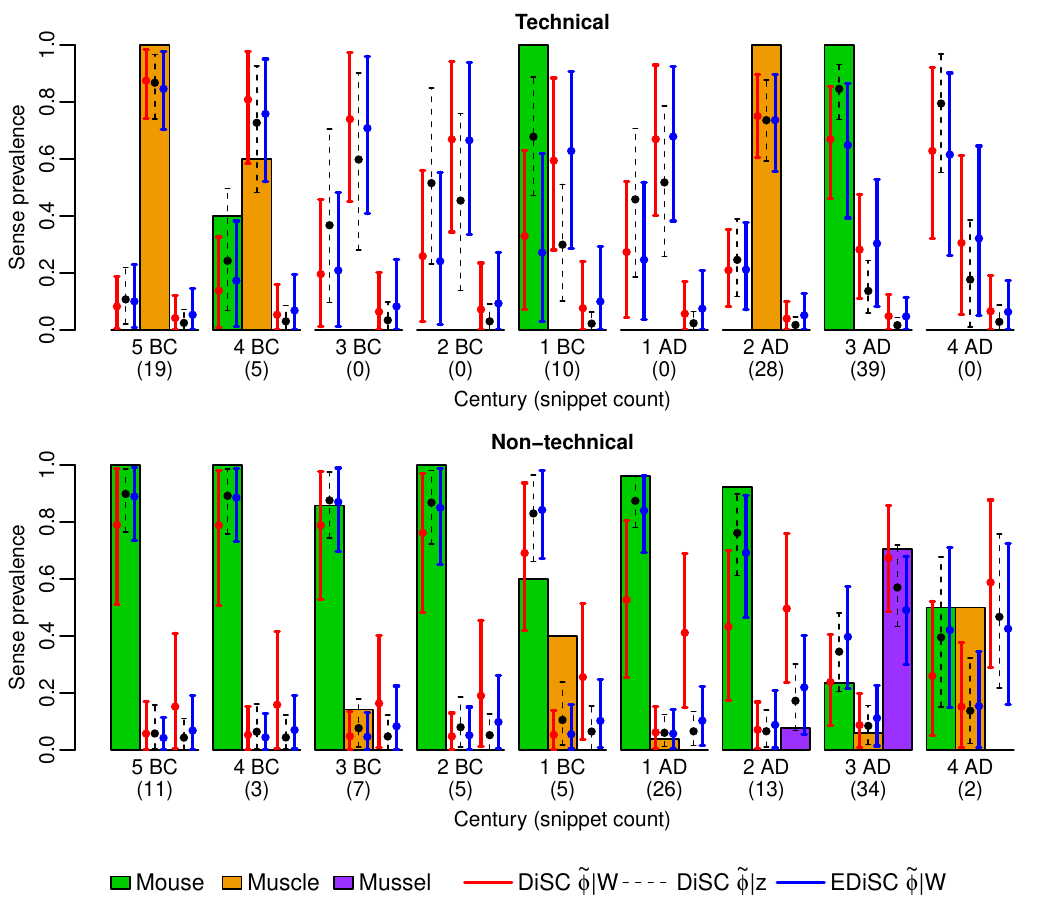}
\caption{``Mus'' expert-annotated empirical sense prevalence (coloured bars), and 95\% HPD intervals (error bars) and posterior means (circles) from the model output}
\label{fig:mus_phi_error_bars}
\end{figure}

\begin{table}[!t]
\caption{\label{tab:BF_mus}``Mus'' Bayes factors $\text{BF}_{01}$ on a $\log_{10}$ scale for nested `true' model $H_0: \tilde{\phi}^{g,t} \in \mathcal{S}^{g,t}$ over each of $H_1:$ DiSC and $H_1:$ EDiSC}
\centering
\begin{tabular}{ l l rrrrrrrrr}
    \toprule
    Model & Genre & 5 BC & 4 BC & 3 BC & 2 BC & 1 BC & 1 AD & 2 AD & 3 AD & 4 AD \\
    \midrule
    DiSC  & technical & 1.37 & 0.98 & 0.80 & 0.61 & 0.46 & 1.00 & 1.56 & 1.12 & 0.85 \\
    EDiSC & technical & 1.30 & 0.98 & 0.77 & 0.51 & 0.32 & 0.85 & 1.44 & 1.10 & 0.85 \\
    \midrule
    DiSC  & non-tech & 0.97 & 0.97 & 0.85 & 0.75 & 0.36 & $-0.36$ & 0.01 & 0.83 & 0.39 \\
    EDiSC & non-tech & 1.20 & 1.20 & 1.09 & 1.05 & 0.89 & 1.24 & 0.99 & 0.90 & 0.48 \\
    \bottomrule
\end{tabular}
\end{table}

Following on from Section~\ref{sec:sense_prevalence_estimation}, we show the sense prevalence graphs for ``mus'' in Figure~\ref{fig:mus_phi_error_bars} and the Bayes factors in Table~\ref{tab:BF_mus}. As in the ``kosmos'' test case, the EDiSC posterior on the unlabelled data generally matches the ground truth using the labelled data better than DiSC (in 56\% of cases), and also gives more precise credible sets for the non-technical genre. Both models pick up on the emergence of the ``mussel'' sense of ``mus'' in the non-technical genre in 2nd~century~AD, with EDiSC providing more accurate prevalence estimates. As for "kosmos", when EDiSC outperforms DiSC, it does so much more substantially (e.g. in the 1st and 2nd centuries AD for the non-technical genre) than the converse.

\begin{figure}[!t]
\centering
\includegraphics[width = 0.9\textwidth, keepaspectratio]{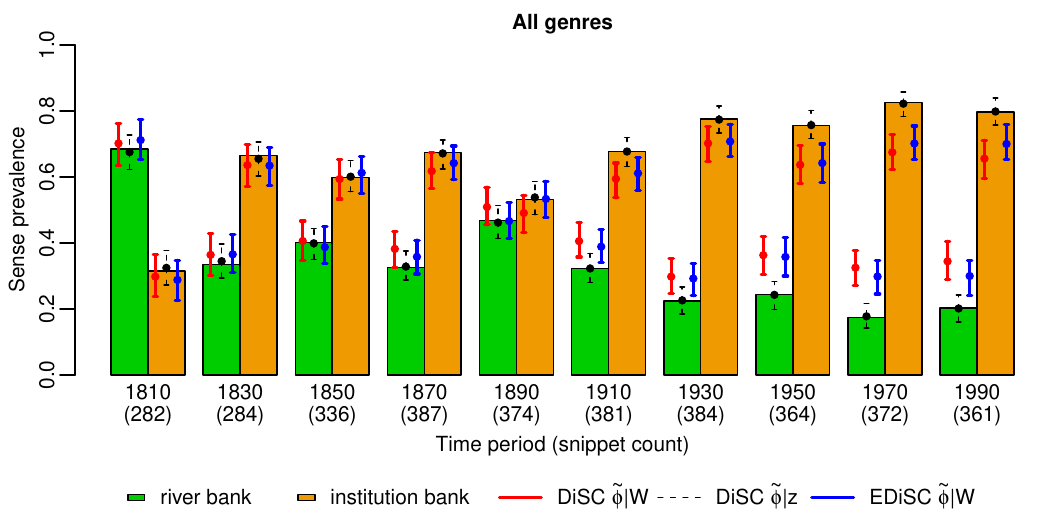}
\caption{``Bank'' manually annotated empirical sense prevalence (coloured bars), and 95\% HPD intervals (error bars) and posterior means (circles) from the model output}
\label{fig:bank_phi_error_bars}
\end{figure}

\begin{table}[!t]
\caption{\label{tab:BF_bank}``Bank'' Bayes factors $\text{BF}_{01}$ on a $\log_{10}$ scale for nested `true' model $H_0: \tilde{\phi}^{g,t} \in \mathcal{S}^{g,t}$ over each of $H_1:$ DiSC and $H_1:$ EDiSC. \textcolor{red}{Red} indicates incorrect rejection of $H_0$.}
\centering
\begin{tabular}{ l rrrrrrrrrr}
    \toprule
    Model & 1810 & 1830 & 1850 & 1870 & 1890 & 1910 & 1930 & 1950 & 1970 & 1990 \\
    \midrule
    DiSC  & 0.85 & 0.91 & 0.98 & 0.67 & 0.73 & $-0.05$ & 0.13 & \textcolor{red}{$-1.60$} & \textcolor{red}{$-\infty$} & \textcolor{red}{$-2.30$} \\
    EDiSC & 0.79 & 0.91 & 0.98 & 0.92 & 0.95 & 0.42 & 0.28 & $-0.91$ & \textcolor{red}{$-1.70$} & $-0.55$ \\   
    \bottomrule
\end{tabular}
\end{table}

The sense prevalence graphs for ``bank'' are given in Figure~\ref{fig:bank_phi_error_bars} and the Bayes factors in Table~\ref{tab:BF_bank}. This is an easy test case where DiSC was already performing well. The improvement provided by EDiSC is therefore only marginal, but it is nevertheless noticeable for the later time periods. EDiSC outperforms DiSC in 90\% of cases, and results in only one incorrect rejection of $H_0$ as opposed to three for DiSC. We see some divergence between the model posteriors and the ground truth in later time periods, resulting in negative Bayes factors. This is because the modelling choice of $K=2$ is a little restrictive. In our experiments with higher $K$ values (which we do not show here), the posteriors recover the ground truth much more closely. However, as discussed in Section~\ref{sec:model_selection}, foreknowledge of the truth cannot be used in model selection. We prioritise interpretability, and the model is performing adequately with our modelling choices. The code is available, and we encourage user exploration.

\begin{figure}[!p]
\centering
\includegraphics[width = 0.9\textwidth, keepaspectratio]{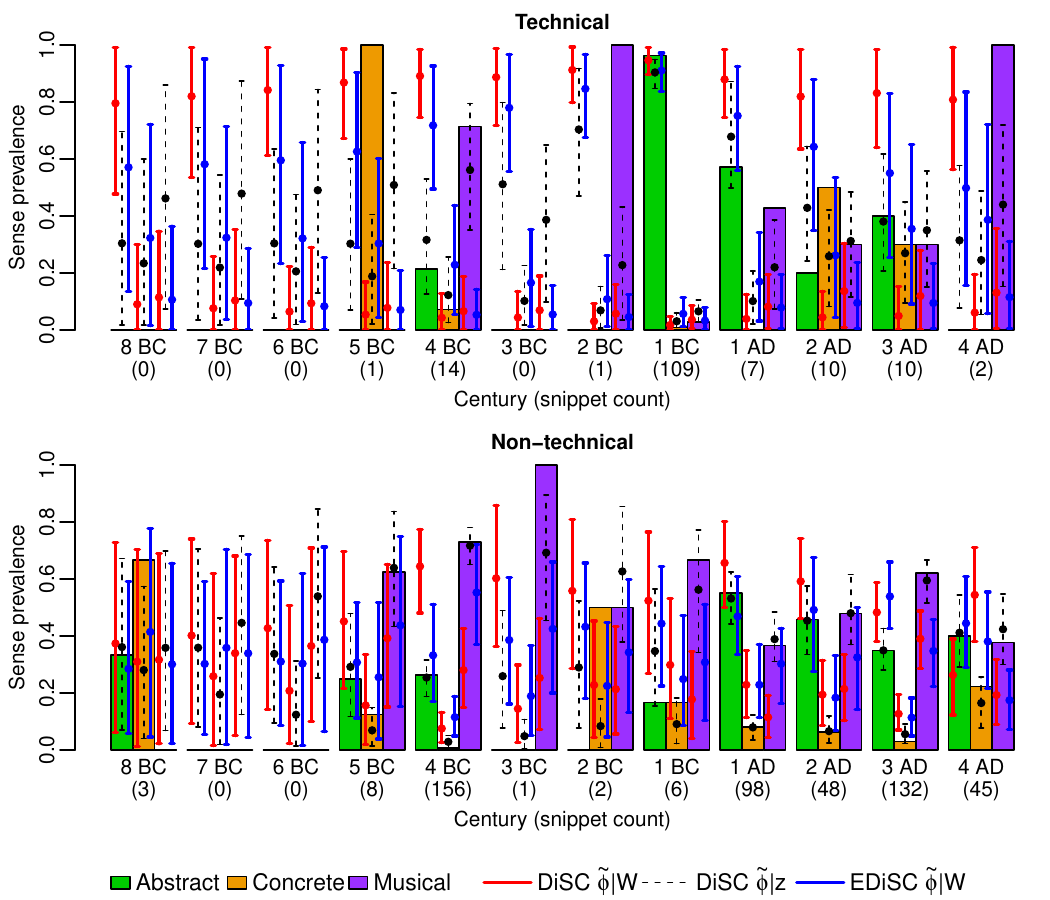}
\caption{``Harmonia'' expert-annotated empirical sense prevalence (coloured bars), and 95\% HPD intervals (error bars) and posterior means (circles) from the model output}
\label{fig:harmonia_phi_error_bars}
\end{figure}

\begin{table}[!p]
\caption{\label{tab:BF_harmonia}``Harmonia'' Bayes factors $\text{BF}_{01}$ on a $\log_{10}$ scale for nested `true' model $H_0: \tilde{\phi}^{g,t} \in \mathcal{S}^{g,t}$ over each of $H_1:$ DiSC and $H_1:$ EDiSC. \textcolor{red}{Red} indicates incorrect rejection of $H_0$.}
\centering
\begin{adjustbox}{max width=\textwidth}
\begin{tabular}{ l l rrrrrrrrrrrr}
    \toprule
    Model & Genre & 8BC & 7BC & 6BC & 5BC & 4BC & 3BC & 2BC & 1BC & 1AD & 2AD & 3AD & 4AD \\
    \midrule
    DiSC  & technical & $-0.60$ & $-0.69$ & \textcolor{red}{$-1.06$} & \textcolor{red}{$-1.38$} & \textcolor{red}{$-\infty$} & $-0.48$ & 0.43 & 1.36 & 0.25 & $-0.95$ & \textcolor{red}{$-1.16$} & \textcolor{red}{$-1.10$} \\
    EDiSC & technical & $-0.27$ & $-0.38$ & $-0.54$ & \textcolor{red}{$-1.04$} & \textcolor{red}{$-2.24$} & $-0.34$ & 0.43 & 1.39 & 0.43 & $-0.05$ & $-0.25$ & $-0.32$ \\
    \midrule
    DiSC  & non-tech & 0.14 & 0.22 & 0.32 & 0.39 & \textcolor{red}{$-\infty$} & $-0.47$ & $-0.46$ & $-0.93$ & \textcolor{red}{$-\infty$} & $-0.96$ & $-0.58$ & \textcolor{red}{$-\infty$} \\
    EDiSC & non-tech & 0.13 & 0.16 & 0.22 & 0.37 & $-0.05$ & 0.17 & 0.11 & 0.05 & 0.12 & 0.39 & $-0.58$ & \textcolor{red}{$-1.01$} \\
    \bottomrule
\end{tabular}
\end{adjustbox}
\end{table}

\begin{table}[!p]
\caption{``Harmonia'' counts and top 10 context words for each expert-annotated sense. Repeated words are shown in \textcolor{red}{red}.} 
\label{tab:harmonia}
\centering
\begin{adjustbox}{max width=\textwidth}
\begin{tabular}{l S[table-format=3.0] p{0.09\linewidth} p{0.09\linewidth} p{0.09\linewidth} p{0.09\linewidth} p{0.09\linewidth} p{0.09\linewidth} p{0.09\linewidth} p{0.09\linewidth} p{0.09\linewidth} p{0.09\linewidth} }
\toprule
Sense & Count & \multicolumn{10}{l}{Top 10 context words under expert sense-annotation} \\
\midrule
abstract & 303 & \textgreek{γίγνομαι} & \textgreek{\textcolor{red}{λόγος}} & \textgreek{\textcolor{red}{πᾶς}} & \textgreek{\textcolor{red}{ποιέω}} & \textgreek{\textcolor{red}{ἔχω}} & \textgreek{ψυχή} & \textgreek{ἁρμονία} & \textgreek{ἀριθμός} & \textgreek{\textcolor{red}{εἷς}} & \textgreek{πολύς} \\
& & \footnotesize (become) & \footnotesize (subject matter) & \footnotesize (all) & \footnotesize (to make) & \footnotesize (to have) & \footnotesize (spirit, soul) & \footnotesize (harmonia) & \footnotesize (number) & \footnotesize (one) & \footnotesize (many, much) \\
concrete & 42 & \textgreek{ὀστέον} & \textgreek{λίθος} & \textgreek{μέγεθος} & \textgreek{ὅσος} & \textgreek{\textcolor{red}{φημί}} & \textgreek{κόσμος} & \textgreek{σῶμα} & \textgreek{μέγας} & \textgreek{οὐκέτι} & \textgreek{\textcolor{red}{πᾶς}} \\
& & \footnotesize (bone, rock) & \footnotesize (stone) & \footnotesize (height) & \footnotesize (how much) & \footnotesize (to speak) & \footnotesize (kosmos) & \footnotesize (body) & \footnotesize (large) & \footnotesize (no more) & \footnotesize (all) \\
musical & 308 & \textgreek{ῥυθμός} & \textgreek{\textcolor{red}{πᾶς}} & \textgreek{\textcolor{red}{ἔχω}} & \textgreek{\textcolor{red}{φημί}} & \textgreek{\textcolor{red}{λόγος}} & \textgreek{πρότερος} & \textgreek{ἁρμονία} & \textgreek{\textcolor{red}{ποιέω}} & \textgreek{\textcolor{red}{εἷς}} & \textgreek{καλέω} \\
& & \footnotesize (rhythm) & \footnotesize (all) & \footnotesize (to have) & \footnotesize (to speak) & \footnotesize (subject matter) & \footnotesize (before) & \footnotesize (harmonia) & \footnotesize (to make) & \footnotesize (one) & \footnotesize (to call, summon) \\
\bottomrule
\end{tabular}
\end{adjustbox}
\end{table}

The sense prevalence for ``harmonia'' is shown in Figure~\ref{fig:harmonia_phi_error_bars} and the Bayes factors in Table~\ref{tab:BF_harmonia}. In contrast to ``bank'', ``harmonia'' is a particularly challenging test case, since the data here is particularly sparse. The ``concrete'' sense of ``harmonia'' is severely under-represented in the data as shown in Table~\ref{tab:harmonia}, and there is a high degree of overlap in probable context words under expert-annotation. The under-represented sense has relatively little effect on the likelihood, so the MCMC targeting the posterior struggles to recognise it. The high overlap in probable words means that the true senses are not very distinct, which understandably affects the models' ability to separate them. Some of the overlapping words appear to be function words based on online translations, so perhaps more targeted data filtering (cf. Section~\ref{sec:data_and_problem}) aided by expert domain-knowledge would be helpful. However, we have not explored this. Despite these challenges, EDiSC performs remarkably well in recovering the truth, as evidenced by the high degree of overlap between the unlabelled and labelled posteriors. EDiSC outperforms DiSC in 83\% of cases, and incorrectly rejects $H_0$ thrice versus eight times for DiSC. DiSC appears to have a strong bias for the ``abstract'' sense in the technical genre, but EDiSC fares much better, benefiting from the extra semantic information contained in the embeddings that is lacking in the sparse snippet data. Accurate sense-change analysis for ``harmonia'' was not possible using DiSC, but is now possible with EDiSC.

\section{MCMC convergence issues} \label{sec:convergence}

As discussed in Sections~\ref{sec:inference} and~\ref{sec:model_selection}, MCMC convergence is essential for inference and model selection. However, getting the MCMC to converge for our test cases with these models is not easy. In this section, we discuss some of the issues encountered and possible strategies for overcoming them. 

\begin{figure}[!t]
\centering
\makebox[\textwidth][c]{\includegraphics[width = 1.1\textwidth, keepaspectratio]{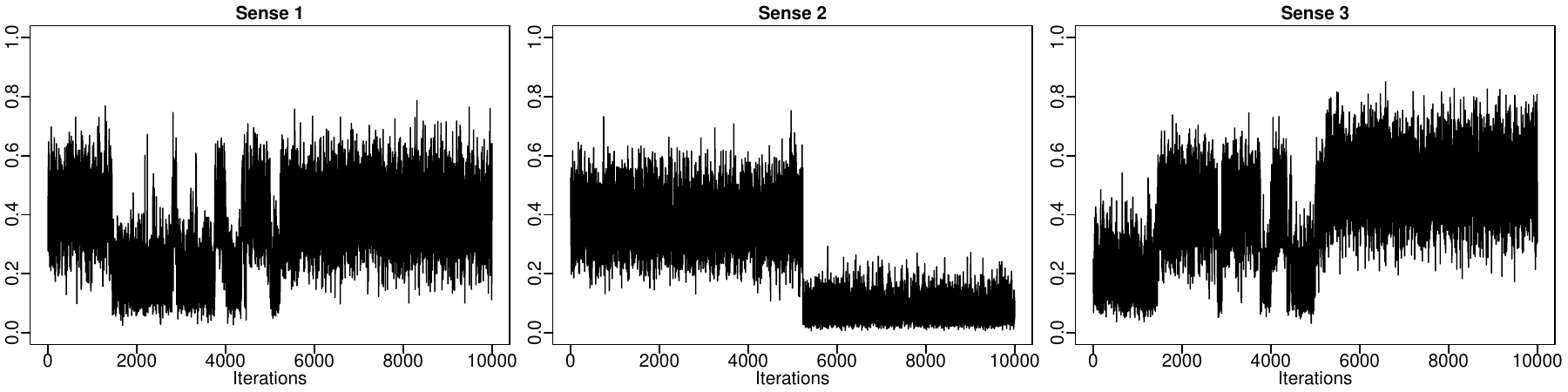}}
\vspace{-20pt}
\caption{$\tilde{\phi}^{g,t}$ trace plots showing the metastable states before MCMC convergence. Here, we run Stan's NUTS on EDiSC for ``mus'' with $M=300,K=3$, and show the plots for $g=2,t=8$.}
\label{fig:mus_metastability_phi}
\end{figure}

\begin{figure}[t]
\centering
\includegraphics[width = 0.75\textwidth, keepaspectratio]{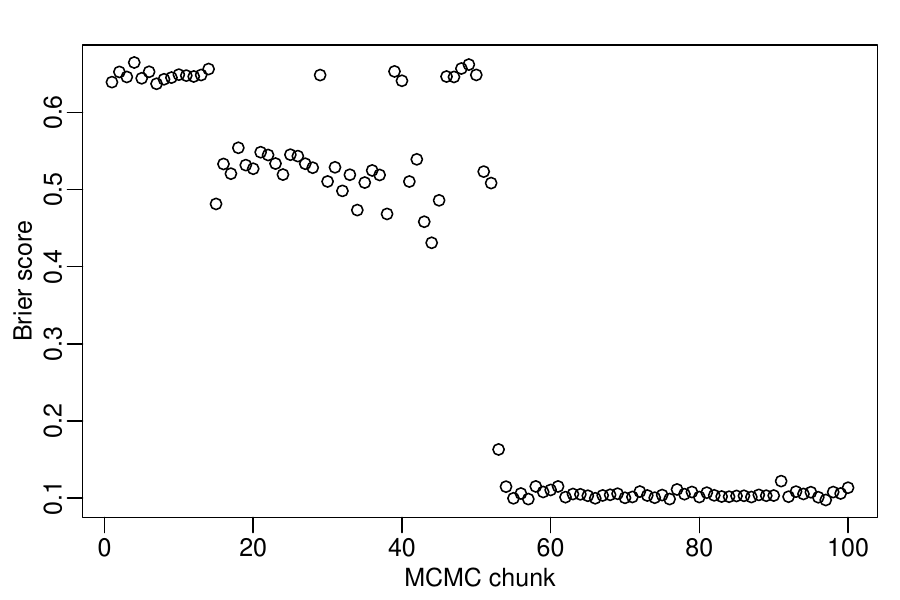}
\caption{Brier scores computed on sequential MCMC chunks of 100 samples each, showing the two metastable states before convergence}
\label{fig:mus_metastability_BS}
\end{figure}

A sampler may sometimes get stuck in a metastable state and fail to converge. Running the sampler for longer often resolves the issue, but this is not guaranteed in practice. As an example, in one of our runs using Stan's NUTS on EDiSC ($M=300, K=3$) for ``mus'', we ran the sampler for 100k iterations, saving every 10th iteration. The sampler was stuck between two different metastable states before eventually finding the correct mode. This can be visualised most easily using the $\tilde{\phi}$ trace plots in Figure~\ref{fig:mus_metastability_phi}. The Brier scores in the metastable states are also shown in Figure~\ref{fig:mus_metastability_BS} to demonstrate the issue, but this cannot be used as a diagnostic in practice since we do not have the ground truth.
Note that this is not a case of label switching: there is no permutation of model senses $1:K$ in either of the metastable states that is equivalent to the posterior in the converged state.

Another technique commonly used in the MCMC literature to overcome metastability is parallel tempering (e.g. the original \citealt{geyer1991markov} or the state of the art \citealt{10.1111/rssb.12464}). This uses MCMC on multiple cores to target, in parallel, a \textit{tempered} or \textit{annealed} posterior in which the likelihood is raised to an inverse `temperature' $\lambda$. Samples are obtained from the `coldest' chain with $\lambda = 1$, which is just the posterior. However, swap moves between chains are proposed, and accepted according to the Hastings ratio, where `hotter' chains use a schedule of $\lambda$ values less than 1, thus encouraging mixing by making the target distribution more diffuse. A problem with this approach in our test cases is that, in order to get reasonable acceptance rates for swap moves between chains, the tempering schedules need to have $\lambda$ values very close to each other. Hence, in order to get mixing at higher temperatures, we need a lot of chains; and this is not achievable with limited computing resources.

However, the tempering or annealing trick \citep{geman1984stochastic, hajek1988cooling} predates parallel tempering, and can be used on a single core during the burn-in phase of the MCMC (and then switched off, so we target the untempered posterior). This allows more mixing earlier in the chain and ensures that, at the least, the sampler does not get stuck in a metastable state due to the starting configuration. In our HMC implementations, we use a simple tempering schedule
\begin{equation} \label{eq:inv_temperature}
    \lambda_n = \lambda_\text{min} + (1 - \lambda_\text{min}) \left( \frac{n}{N_\text{temp}} \right)^\beta \text{,}
\end{equation}
where $n$ is the MCMC iteration number, $N_\text{temp}$ is the number of iterations for which to use tempering, $\lambda_\text{min}$ is the minimum inverse temperature, and $\beta \leq 1$ determines the rate of change in $\lambda_n$ (with $\beta=1$ giving a linear schedule and $\beta<1$ giving a schedule increasing at a decreasing rate). We find that $\lambda_\text{min} = 0.1$, $\beta = 1/3$ and $N_\text{temp}$ set equal to half the total number of MCMC iterations works adequately in our experiments. Also, we temper the likelihood \eqref{eq:likelihood} when targeting $\phi$ and $\chi$ only, since $\theta$ and $\varsigma$ are very well-informed by the data as it is.

\section{Gradient-based MCMC sampling} \label{sec:samplers}

As mentioned in Section~\ref{sec:inference}, for our MALA and HMC implementations of EDiSC, we use analytically derived gradients. We can do MALA or HMC sampling for $\phi,\theta,\chi,\varsigma$ in the same way as discussed in \citet[Appendix~C]{DiSC_https://doi.org/10.1111/rssc.12591}: alternately sample each variable while conditioning on the others. Inference for $\phi$ in EDiSC is unchanged compared to DiSC, whereas to get proposals for $\theta,\chi,\varsigma$, we need to derive the gradients for the log-likelihood 
\begin{equation*}
    \log p(W | \phi, \psi) = \sum_{d=1}^D \log \sum_{k=1}^K \tilde{\phi}_k^{\gamma_d,\tau_d} \prod_{i=i_1}^{i_{L_d}} \tilde{\psi}_{w_{d,i}}^{k,\tau_d}
\end{equation*} 
with respect to these variables, which we do in this section. Finally, we give the HMC algorithm used in our R implementation.

\subsection{Derivation of $\nabla_{\xi^{k,t}} \log p(W_{\mathcal{D}(1:G,t)} | \phi^{\cdot,t}, \psi^{\cdot,t})$} \label{sec:lik_grad_wrt_xi}

From \citet{DiSC_https://doi.org/10.1111/rssc.12591} equation (32) we have 
\begin{equation} \label{eq:diff_log_p_psi_final}
    \frac{\partial}{\partial \psi^{k,t}_v} \log p(W_{\mathcal{D}(1:G,t)} | \phi^{\cdot,t}, \psi^{\cdot,t}) = \sum_{d \in \mathcal{D}(1:G,t)} \frac {\tilde{\phi}_k^{\gamma_d,t} \prod_{i=i_1}^{i_{L_d}} \tilde{\psi}_{w_{d,i}}^{k,t}} {\sum_{l=1}^K \tilde{\phi}_l^{\gamma_d,t} \prod_{i=i_1}^{i_{L_d}} \tilde{\psi}_{w_{d,i}}^{l,t}} \left( \sum_{i=i_1}^{i_{L_d}} \mathbb{I}(v=w_{d,i}) - L_d \tilde{\psi}_{v}^{k,t} \right) 
\end{equation}
where $\mathcal{D}(g,t) = \{ d: \gamma_d \in g \text{ and } \tau_d \in t \}$ is the set of snippet indices for genre(s) $g$ and time(s) $t$. Also, by the chain rule we have
\begin{equation*} \label{eq:diff_log_p_xi}
    \frac{\partial}{\partial \xi^{k,t}_j} \log p(W_{\mathcal{D}(1:G,t)} | \phi^{\cdot,t}, \psi^{\cdot,t}) = \sum_{v=1}^V \frac{\partial}{\partial  \psi^{k,t}_v} \log p(W_{\mathcal{D}(1:G,t)} | \phi^{\cdot,t}, \psi^{\cdot,t}) \frac{\partial \psi^{k,t}_v}{\partial  \xi^{k,t}_j} \text{.}
\end{equation*}

Now $\psi^{k,t}_v = \rho_v^\text{T} \xi^{k,t} + \varsigma_v$ gives $\frac{\partial \psi^{k,t}_v}{\partial \xi^{k,t}_j} = \rho_{v,j}$, and so we have
\begin{align}
     \frac{\partial}{\partial \xi^{k,t}_j} \log p(W_{\mathcal{D}(1:G,t)} | \phi^{\cdot,t}, \psi^{\cdot,t}) 
     &= \sum_{v=1}^V \rho_{v,j} \frac{\partial}{\partial  \psi^{k,t}_v} \log p(W_{\mathcal{D}(1:G,t)} | \phi^{\cdot,t}, \psi^{\cdot,t}) \nonumber \\
     &= \sum_{d \in \mathcal{D}(1:G,t)} \frac {\tilde{\phi}_k^{\gamma_d,t} \prod_{i=i_1}^{i_{L_d}} \tilde{\psi}_{w_{d,i}}^{k,t}} {\sum_{l=1}^K \tilde{\phi}_l^{\gamma_d,t} \prod_{i=i_1}^{i_{L_d}} \tilde{\psi}_{w_{d,i}}^{l,t}} \left( \sum_{i=i_1}^{i_{L_d}} \sum_{v=1}^V \rho_{v,j} \mathbb{I}(v=w_{d,i}) - L_d \sum_{v=1}^V \rho_{v,j} \tilde{\psi}_{v}^{k,t} \right) \nonumber \\
     &= \sum_{d \in \mathcal{D}(1:G,t)} \frac {\tilde{\phi}_k^{\gamma_d,t} \prod_{i=i_1}^{i_{L_d}} \tilde{\psi}_{w_{d,i}}^{k,t}} {\sum_{l=1}^K \tilde{\phi}_l^{\gamma_d,t} \prod_{i=i_1}^{i_{L_d}} \tilde{\psi}_{w_{d,i}}^{l,t}} \left( \sum_{i=i_1}^{i_{L_d}} \rho_{w_{d,i},j} - L_d \rho_{\cdot,j}^\text{T} \tilde{\psi}^{k,t} \right) \label{eq:diff_log_p_xi_final}
\end{align}
which are the elements of vector $\nabla_{\xi^{k,t}} \log p(W_{\mathcal{D}(1:G,t)} | \phi^{\cdot,t}, \psi^{\cdot,t})$ for $j \in \{1,\dots,M\}$.

\subsection{Derivation of $\nabla_{\theta^{t}} \log p(W_{\mathcal{D}(1:G,t)} | \phi^{\cdot,t}, \psi^{\cdot,t})$}

The relationship $\xi^{k,t}_j = \chi^k_j + \theta^t_j$ gives $\frac{\partial \xi^{k,t}_j}{\partial \theta^t_j} = 1$ for all $k \in \{1,\dots,K\}$, so applying the chain rule to \eqref{eq:diff_log_p_xi_final} we get
\begin{align*}
    \frac{\partial}{\partial \theta^t_j} \log p(W_{\mathcal{D}(1:G,t)} | \phi^{\cdot,t}, \psi^{\cdot,t})
    &= \sum_{k=1}^K \frac{\partial}{\partial \xi^{k,t}_j} \log p(W_{\mathcal{D}(1:G,t)} | \phi^{\cdot,t}, \psi^{\cdot,t})  \\
    &= \sum_{d \in \mathcal{D}(1:G,t)} \left( \sum_{i=i_1}^{i_{L_d}} \rho_{w_{d,i},j} - L_d \sum_{k=1}^K \frac {\tilde{\phi}_k^{\gamma_d,t} \prod_{i=i_1}^{i_{L_d}} \tilde{\psi}_{w_{d,i}}^{k,t}} {\sum_{l=1}^K \tilde{\phi}_l^{\gamma_d,t} \prod_{i=i_1}^{i_{L_d}} \tilde{\psi}_{w_{d,i}}^{l,t}} \rho_{\cdot,j}^\text{T} \tilde{\psi}^{k,t} \right)
\end{align*}
which are the elements of vector $\nabla_{\theta^{t}} \log p(W_{\mathcal{D}(1:G,t)} | \phi^{\cdot,t}, \psi^{\cdot,t})$ for $j \in \{1,\dots,M\}$.

\subsection{Derivation of $\nabla_{\chi^k} \log p(W | \phi, \psi)$}

The relationship $\xi^{k,t}_j = \chi^k_j + \theta^t_j$ gives $\frac{\partial \xi^{k,t}_j}{\partial \chi^k_j} = 1$ for all $t \in \{1,\dots,T\}$, so given the independence between time periods and applying the chain rule to \eqref{eq:diff_log_p_xi_final} we get
\begin{align*}
     \frac{\partial}{\partial \chi^k_j} \log p(W | \phi, \psi) 
     &= \sum_{t=1}^T \frac{\partial}{\partial \xi^{k,t}_j} \log p(W_{\mathcal{D}(1:G,t)} | \phi^{\cdot,t}, \psi^{\cdot,t}) \\
     &= \sum_{d=1}^D \frac { \tilde{\phi}_k^{\gamma_d,\tau_d} \prod_{i=i_1}^{i_{L_d}} \tilde{\psi}_{w_{d,i}}^{k,\tau_d} } {\sum_{l=1}^K \tilde{\phi}_l^{\gamma_d,\tau_d} \prod_{i=i_1}^{i_{L_d}} \tilde{\psi}_{w_{d,i}}^{l,\tau_d}} \left( \sum_{i=i_1}^{i_{L_d}} \rho_{w_{d,i},j} - L_d \rho_{\cdot,j}^\text{T} \tilde{\psi}^{k,\tau_d} \right)
\end{align*}
which are the elements of vector $\nabla_{\chi^k} \log p(W | \phi, \psi)$ for $j \in \{1,\dots,M\}$.

\subsection{Derivation of $\nabla_{\varsigma} \log p(W | \phi, \psi)$}

The relationship $\psi^{k,t}_j = \rho_j^\text{T} \xi^{k,t} + \varsigma_j$ gives $\frac{\partial \psi^{k,t}_j}{\partial \varsigma_j} = 1$ for all $k \in \{1,\dots,K\}$ and for all $t \in \{1,\dots,T\}$, so given the independence between time periods and applying the chain rule to \eqref{eq:diff_log_p_psi_final} we get 
\begin{align} 
    \frac{\partial}{\partial \varsigma_j} \log p(W | \phi, \psi) 
    &= \sum_{t=1}^T \sum_{k=1}^K \frac{\partial}{\partial \psi^{k,t}_j} \log p(W_{\mathcal{D}(1:G,t)} | \phi^{\cdot,t}, \psi^{\cdot,t}) \nonumber \\ 
    &= \sum_{d=1}^D \sum_{k=1}^K \frac {\tilde{\phi}_k^{\gamma_d,\tau_d} \prod_{i=i_1}^{i_{L_d}} \tilde{\psi}_{w_{d,i}}^{k,\tau_d}} {\sum_{l=1}^K \tilde{\phi}_l^{\gamma_d,\tau_d} \prod_{i=i_1}^{i_{L_d}} \tilde{\psi}_{w_{d,i}}^{l,\tau_t}} \left( \sum_{i=i_1}^{i_{L_d}} \mathbb{I}(j=w_{d,i}) - L_d \tilde{\psi}_{j}^{k,\tau_d} \right) \nonumber \\
    &= \sum_{d=1}^D \left( \sum_{i=i_1}^{i_{L_d}} \mathbb{I}(j=w_{d,i}) - L_d \sum_{k=1}^K \frac {\tilde{\phi}_k^{\gamma_d,\tau_d} \prod_{i=i_1}^{i_{L_d}} \tilde{\psi}_{w_{d,i}}^{k,\tau_d}} {\sum_{l=1}^K \tilde{\phi}_l^{\gamma_d,\tau_d} \prod_{i=i_1}^{i_{L_d}} \tilde{\psi}_{w_{d,i}}^{l,\tau_t}} \tilde{\psi}_{j}^{k,\tau_d} \right) \label{eq:diff_log_p_sigma_final}
\end{align}
which are the elements of vector $\nabla_{\varsigma} \log p(W | \phi, \psi)$ for $j \in \{1,\dots,V\}$. Note that the term on the left in \eqref{eq:diff_log_p_sigma_final} can be simplified by considering that $\sum_{d=1}^D \sum_{i=i_1}^{i_{L_d}} \mathbb{I}(j=w_{d,i}) = N^{W,\cdot}_{j,\cdot,\cdot}$ is the number of times word $j$ appears in snippets $W$ across all senses and time periods.

\subsection{HMC sampling scheme}

\begin{algorithm}[!t]
\caption{Hamiltonian Monte Carlo (HMC) sampling for EDiSC}
\label{alg:HMC}
\begin{algorithmic}[1]
\RaggedRight

\State set number of MCMC iterations $N$, tempering parameter $N_\text{temp}$, tuning parameters $N_\text{tune}, N_\text{stop}$, and target acceptance rate $\alpha^\text{opt}$
\For {each variable $x \in \{\chi, \theta^t, \phi^{g,t}, \varsigma | g=1,\dots,G; t=1,\dots,T, \}$}
    \State set number of leapfrog steps $LF_x$ and initial proposal scale $\sigma^2_x$
    \State set tempering for variable $x$ on or off
\EndFor

\State initialise $\phi, \chi, \theta$ randomly and $\varsigma=\mathbf{0}$

\For {iteration $n \in 1:N$}    
    \For {each $x \in \{\chi, \theta^t, \phi^{g,t}, \varsigma | g=1,\dots,G; t=1,\dots,T, \}$}
        \State \textbf{if} tempering $x$ and $n \leq N_\text{temp}$ \textbf{then} compute $\lambda_n$ using \eqref{eq:inv_temperature}; \textbf{else} set $\lambda_n=1$
        \State draw initial momentum vector $q \sim \mathcal{N}(\mathbf{0},\mathbf{I})$ of same dimension as $x$
        \State compute initial potential energy $PE_0 = PE(x) = -\log \left( \pi(x) p(W | \phi(x), \psi(x))^{\lambda_n} \right)$ 
        \State compute initial kinetic energy $KE_0 = KE(q) = \frac{1}{2} \sum q^2$ 
        \State make a half step for momentum at the beginning $q \leftarrow q - \frac{1}{2} \sigma_x \nabla_x PE(x)$
        \State save initial $x_0 = x$
        \For {$l \in 1:LF_x$}
            \State make a full step for position $x \leftarrow x + \sigma_x q$
            \State if $l \neq LF_x$, make a full step for momentum $q \leftarrow q - \sigma_x \nabla_x PE(x)$
        \EndFor
        \State save final $x_1 = x$
        \State make a half step for momentum at the end $q \leftarrow q - \frac{1}{2} \sigma_x \nabla_x PE(x)$
        \State compute final potential and kinetic energies $PE_1 = PE(x)$ and $KE_1 = KE(q)$
        \State compute Hastings ratio $\alpha = \min \{1, \exp(PE_0 + KE_0 - PE_1 - KE_1) \}$
        \State with probability $\alpha$, set $x = x_1$ (accept); else set $x = x_0$ (reject)

        \If {$n \geq N_\text{tune}$ and $n \leq N_\text{stop}$} 
            \State compute running acceptance rate $\Bar{\alpha} = \frac{\text{\# accepts}}{N_\text{tune}}$ using last $N_\text{tune}$ iterations
            \State update proposal scale via $\log \sigma_x^2 \leftarrow \log \sigma_x^2 + C_n (\Bar{\alpha} - \alpha^\text{opt})$ with $C_n = \left( \frac{n+1}{N_\text{tune}} \right)^{-0.8}$
        \EndIf
    \EndFor    
\EndFor

\end{algorithmic}
\end{algorithm}

The HMC algorithm used in our implementation is adapted from \citet{2012arXiv1206.1901N}, and is presented within the context of our EDiSC model and sampling scheme in Algorithm~\ref{alg:HMC}. We target one variable at a time, conditioning on all other variables, at the granularity of $\chi, \theta^t, \phi^{g,t}, \varsigma$, in that order. For increased efficiency, we iterate over time $t \in 1:T$ going forward for odd iterations and backward for even iterations. We use 10 leapfrog steps for $\chi$, as this is arguably the most important and challenging variable to sample, and 5 leapfrog steps for all other variables. We run the MCMC for $N$ iterations, which varies considerably between at least 1.5k for "bank" and up to 100k for "mus" in some runs, though typically 3k--10k is sufficient to ensure convergence in our experiments. Gradients for the potential energy $PE(x) = -\log \left( \pi(x) p(W | \phi(x), \psi(x))^{\lambda_n} \right)$ for each variable $x$ take the form $\nabla_x PE(x) = - \nabla_x \log \pi(x) - \lambda_n \nabla_x \log p(W | \phi, \psi)$, where $\nabla_x \log \pi(x)$ are straightforward to compute since the prior densities $\pi(x)$ are normal (see \citealt{DiSC_https://doi.org/10.1111/rssc.12591} Appendix C for explicit expressions), and $\nabla_x \log p(W | \phi, \psi)$ have been derived above.

The tuning scheme used to target an optimal acceptance rate is taken from \citet{shaby2010exploring}. We target an optimal acceptance rate $\alpha^\text{opt} = 0.651$ for HMC as recommended by \citet{beskos2013}. For the initial proposal scales, we slightly adapt the authors' recommendations and use $\sigma^2_\phi = 2.4^2 / (K \times LF_\phi)$, $\sigma^2_\chi = 2.4^2 / ( (MK)^2 \times LF_\chi)$, $\sigma^2_\theta = 2.4^2 / (M^2 \times LF_\theta)$ and $\sigma^2_\varsigma = 2.4^2 / (V \times LF_\varsigma)$, where $LF_x$ is the number of leapfrog steps used for variable $x$. These work adequately for our data, but there is no particular reason to stick with these choices. The main consideration is to strike a balance, via trial and error, between setting the initial proposal scales too large (leading to numerical over/underflow) and too small (leading to slower mixing at the start of the chain). The scales are updated via $\log \sigma_x^2 \leftarrow \log \sigma_x^2 + C_n (\Bar{\alpha} - \alpha^\text{opt})$ at MCMC iteration $n$, using a running acceptance rate $\Bar{\alpha}$ computed on the last $N_\text{tune} = 10$ iterations. We typically stop tuning after $N_\text{stop} = N/2$ iterations, though it is harmless to continue tuning since $C_n = \left( \frac{n+1}{N_\text{tune}} \right)^{-0.8}$ by design converges to zero as $n \rightarrow \infty$ and the tuning becomes minuscule.

\bibliography{bibliography.bib}

\begin{thebibliography}{63}
\providecommand{\natexlab}[1]{#1}
\providecommand{\url}[1]{\texttt{#1}}
\expandafter\ifx\csname urlstyle\endcsname\relax
  \providecommand{\doi}[1]{doi: #1}\else
  \providecommand{\doi}{doi: \begingroup \urlstyle{rm}\Url}\fi

\bibitem[Apidianaki(2022)]{10.1162/coli_a_00474}
{\rm Apidianaki, M.} (2022).
\newblock {From Word Types to Tokens and Back: A Survey of Approaches to Word Meaning Representation and Interpretation}.
\newblock \emph{Computational Linguistics}, {\bf 49}\penalty0 (2), \penalty0 465--523.
\newblock ISSN 0891-2017.
\newblock \doi{10.1162/coli_a_00474}.
\newblock URL \url{https://doi.org/10.1162/coli\_a\_00474}.

\bibitem[Benoit et~al.(2020)Benoit, Muhr, and Watanabe]{stopwords2020}
{\rm Benoit, K., Muhr, D., {\rm and} Watanabe, K.} (2020).
\newblock \emph{Stopwords: Multilingual Stopword Lists}.
\newblock URL \url{https://CRAN.R-project.org/package=stopwords}.
\newblock R package version 2.1.

\bibitem[Berra(2018)]{greek_stopwords}
{\rm Berra, A.} (2018).
\newblock \emph{{A}ncient {G}reek and {L}atin Stopwords for Textual Analysis}.
\newblock URL \url{https://github.com/aurelberra/stopwords}.
\newblock Greek v2.7 as of 30 Oct 2018.

\bibitem[Beskos et~al.(2013)Beskos, Pillai, Roberts, Sanz-Serna, and Stuart]{beskos2013}
{\rm Beskos, A., Pillai, N., Roberts, G., Sanz-Serna, J.-M., {\rm and} Stuart, A.} (2013).
\newblock Optimal tuning of the hybrid {M}onte {C}arlo algorithm.
\newblock \emph{Bernoulli}, {\bf 19}\penalty0 (5A), \penalty0 1501--1534.
\newblock \doi{10.3150/12-BEJ414}.

\bibitem[Bevilacqua et~al.(2021)Bevilacqua, Pasini, Raganato, and Navigli]{bevilacqua2021recent}
{\rm Bevilacqua, M., Pasini, T., Raganato, A., {\rm and} Navigli, R.} (2021).
\newblock Recent trends in word sense disambiguation: A survey.
\newblock In \emph{Proceedings of the Thirtieth International Joint Conference on Artificial Intelligence, IJCAI-21}. International Joint Conference on Artificial Intelligence, Inc.

\bibitem[Blei(2012)]{blei2012probabilistic}
{\rm Blei, D.~M.} (2012).
\newblock Probabilistic topic models.
\newblock \emph{Communications of the ACM}, {\bf 55}\penalty0 (4), \penalty0 77--84.

\bibitem[Blei and Lafferty(2006)]{Blei:2006:DTM:1143844.1143859}
{\rm Blei, D.~M. {\rm and} Lafferty, J.~D.} (2006).
\newblock Dynamic topic models.
\newblock In \emph{Proceedings of the 23rd international conference on Machine learning}, pages 113--120.

\bibitem[Blei et~al.(2003)Blei, Ng, and Jordan]{blei2003latent}
{\rm Blei, D.~M., Ng, A.~Y., {\rm and} Jordan, M.~I.} (2003).
\newblock Latent {D}irichlet {A}llocation.
\newblock \emph{Journal of machine Learning research}, {\bf 3}\penalty0 (Jan), \penalty0 993--1022.

\bibitem[Bojanowski et~al.(2017)Bojanowski, Grave, Joulin, and Mikolov]{10.1162/tacl_a_00051}
{\rm Bojanowski, P., Grave, E., Joulin, A., {\rm and} Mikolov, T.} (2017).
\newblock {Enriching Word Vectors with Subword Information}.
\newblock \emph{Transactions of the Association for Computational Linguistics}, {\bf 5}, \penalty0 135--146.
\newblock ISSN 2307-387X.
\newblock \doi{10.1162/tacl_a_00051}.
\newblock URL \url{https://doi.org/10.1162/tacl\_a\_00051}.

\bibitem[Churchill and Singh(2022)]{churchill2022evolution}
{\rm Churchill, R. {\rm and} Singh, L.} (2022).
\newblock The evolution of topic modeling.
\newblock \emph{ACM Computing Surveys}, {\bf 54}\penalty0 (10s), \penalty0 1--35.

\bibitem[Davies(2010)]{davies2010corpus}
{\rm Davies, M.} (2010).
\newblock The corpus of historical {A}merican {E}nglish: 400 million words, 1810-2009.

\bibitem[Devlin et~al.(2019)Devlin, Chang, Lee, and Toutanova]{devlin-etal-2019-bert}
{\rm Devlin, J., Chang, M.-W., Lee, K., {\rm and} Toutanova, K.} (2019).
\newblock {BERT}: Pre-training of deep bidirectional transformers for language understanding.
\newblock In \emph{Proceedings of the 2019 Conference of the North {A}merican Chapter of the Association for Computational Linguistics: Human Language Technologies, Volume 1 (Long and Short Papers)}, pages 4171--4186, Minneapolis, Minnesota, June 2019. Association for Computational Linguistics.
\newblock \doi{10.18653/v1/N19-1423}.
\newblock URL \url{https://aclanthology.org/N19-1423}.

\bibitem[Dieng et~al.(2019)Dieng, Ruiz, and Blei]{dieng2019dynamic}
{\rm Dieng, A.~B., Ruiz, F.~J., {\rm and} Blei, D.~M.} (2019).
\newblock The dynamic embedded topic model.
\newblock \emph{arXiv preprint arXiv:1907.05545}.

\bibitem[Dieng et~al.(2020)Dieng, Ruiz, and Blei]{10.1162/tacl_a_00325}
{\rm Dieng, A.~B., Ruiz, F. J.~R., {\rm and} Blei, D.~M.} (2020).
\newblock {Topic Modeling in Embedding Spaces}.
\newblock \emph{Transactions of the Association for Computational Linguistics}, {\bf 8}, \penalty0 439--453.
\newblock ISSN 2307-387X.
\newblock \doi{10.1162/tacl_a_00325}.
\newblock URL \url{https://doi.org/10.1162/tacl\_a\_00325}.

\bibitem[{Duane} et~al.(1987){Duane}, {Kennedy}, {Pendleton}, and {Roweth}]{1987PhLB..195..216D}
{\rm {Duane}, S., {Kennedy}, A.~D., {Pendleton}, B.~J., {\rm and} {Roweth}, D.} (1987).
\newblock {Hybrid {M}onte {C}arlo}.
\newblock \emph{Physics Letters B}, {\bf 195}\penalty0 (2), \penalty0 216--222.
\newblock \doi{10.1016/0370-2693(87)91197-X}.

\bibitem[{Dubossarsky} et~al.(2019){Dubossarsky}, {Hengchen}, {Tahmasebi}, and {Schlechtweg}]{2019arXiv190601688D}
{\rm {Dubossarsky}, H., {Hengchen}, S., {Tahmasebi}, N., {\rm and} {Schlechtweg}, D.} (2019).
\newblock {Time-Out: Temporal Referencing for Robust Modeling of Lexical Semantic Change}.
\newblock \emph{arXiv e-prints}, art. arXiv:1906.01688.

\bibitem[Frermann and Lapata(2016)]{frermann2016bayesian}
{\rm Frermann, L. {\rm and} Lapata, M.} (2016).
\newblock A {B}ayesian model of diachronic meaning change.
\newblock \emph{Transactions of the Association for Computational Linguistics}, {\bf 4}, \penalty0 31--45.

\bibitem[Geman and Geman(1984)]{geman1984stochastic}
{\rm Geman, S. {\rm and} Geman, D.} (1984).
\newblock Stochastic relaxation, {G}ibbs distributions, and the {B}ayesian restoration of images.
\newblock \emph{IEEE Transactions on pattern analysis and machine intelligence}, \penalty0 (6), \penalty0 721--741.

\bibitem[Geyer et~al.(1991)Geyer, Keramidas, and Kaufman]{geyer1991markov}
{\rm Geyer, C.~J., Keramidas, E., {\rm and} Kaufman, S.} (1991).
\newblock Markov {C}hain {M}onte {C}arlo {M}aximum {L}ikelihood. {I}nterface {F}oundation of {N}orth {A}merica.

\bibitem[Hajek(1988)]{hajek1988cooling}
{\rm Hajek, B.} (1988).
\newblock Cooling schedules for optimal annealing.
\newblock \emph{Mathematics of operations research}, {\bf 13}\penalty0 (2), \penalty0 311--329.

\bibitem[{Hamilton} et~al.(2016){Hamilton}, {Leskovec}, and {Jurafsky}]{2016arXiv160509096H}
{\rm {Hamilton}, W.~L., {Leskovec}, J., {\rm and} {Jurafsky}, D.} (2016).
\newblock {Diachronic Word Embeddings Reveal Statistical Laws of Semantic Change}.
\newblock \emph{arXiv e-prints}, art. arXiv:1605.09096.

\bibitem[Hoffman and Gelman(2014)]{hoffman2014no}
{\rm Hoffman, M.~D. {\rm and} Gelman, A.} (2014).
\newblock The {N}o-{U}-{T}urn sampler: adaptively setting path lengths in {H}amiltonian {M}onte {C}arlo.
\newblock \emph{J. Mach. Learn. Res.}, {\bf 15}\penalty0 (1), \penalty0 1593--1623.

\bibitem[Kass and Raftery(1995)]{doi:10.1080/01621459.1995.10476572}
{\rm Kass, R.~E. {\rm and} Raftery, A.~E.} (1995).
\newblock {B}ayes factors.
\newblock \emph{Journal of the {A}merican {S}tatistical {A}ssociation}, {\bf 90}\penalty0 (430), \penalty0 773--795.
\newblock \doi{10.1080/01621459.1995.10476572}.

\bibitem[{Kucukelbir} et~al.(2015){Kucukelbir}, {Ranganath}, {Gelman}, and {Blei}]{2015arXiv150603431K}
{\rm {Kucukelbir}, A., {Ranganath}, R., {Gelman}, A., {\rm and} {Blei}, D.~M.} (2015).
\newblock {Automatic Variational Inference in Stan}.
\newblock \emph{arXiv e-prints}, art. arXiv:1506.03431.
\newblock \doi{10.48550/arXiv.1506.03431}.

\bibitem[Kulkarni et~al.(2015)Kulkarni, Al-Rfou, Perozzi, and Skiena]{Kulkarni:2015:SSD:2736277.2741627}
{\rm Kulkarni, V., Al-Rfou, R., Perozzi, B., {\rm and} Skiena, S.} (2015).
\newblock Statistically significant detection of linguistic change.
\newblock In \emph{Proceedings of the 24th International Conference on World Wide Web}, pages 625--635.

\bibitem[Kutuzov et~al.(2018)Kutuzov, {\O}vrelid, Szymanski, and Velldal]{kutuzov-etal-2018-diachronic}
{\rm Kutuzov, A., {\O}vrelid, L., Szymanski, T., {\rm and} Velldal, E.} (2018).
\newblock Diachronic word embeddings and semantic shifts: a survey.
\newblock In \emph{Proceedings of the 27th International Conference on Computational Linguistics}, pages 1384--1397, Santa Fe, New Mexico, USA, August 2018. Association for Computational Linguistics.
\newblock URL \url{https://aclanthology.org/C18-1117}.

\bibitem[McGillivray et~al.(2019)McGillivray, Hengchen, Lähteenoja, Palma, and Vatri]{10.1093/llc/fqz036}
{\rm McGillivray, B., Hengchen, S., Lähteenoja, V., Palma, M., {\rm and} Vatri, A.} (2019).
\newblock {A computational approach to lexical polysemy in {A}ncient {G}reek}.
\newblock \emph{Digital Scholarship in the Humanities}, {\bf 34}\penalty0 (4), \penalty0 893--907.
\newblock ISSN 2055-7671.
\newblock \doi{10.1093/llc/fqz036}.
\newblock URL \url{https://doi.org/10.1093/llc/fqz036}.

\bibitem[Melamud et~al.(2016)Melamud, Goldberger, and Dagan]{melamud2016context2vec}
{\rm Melamud, O., Goldberger, J., {\rm and} Dagan, I.} (2016).
\newblock context2vec: Learning generic context embedding with bidirectional {LSTM}.
\newblock In \emph{Proceedings of the 20th SIGNLL conference on computational natural language learning}, pages 51--61.

\bibitem[Mikolov et~al.(2013{\natexlab{a}})Mikolov, Chen, Corrado, and Dean]{2013arXiv1301.3781M}
{\rm Mikolov, T., Chen, K., Corrado, G., {\rm and} Dean, J.} (2013{\natexlab{a}}).
\newblock {Efficient Estimation of Word Representations in Vector Space}.
\newblock \emph{arXiv e-prints}, art. arXiv:1301.3781.

\bibitem[Mikolov et~al.(2013{\natexlab{b}})Mikolov, Sutskever, Chen, Corrado, and Dean]{NIPS2013_5021}
{\rm Mikolov, T., Sutskever, I., Chen, K., Corrado, G.~S., {\rm and} Dean, J.} (2013{\natexlab{b}}).
\newblock Distributed representations of words and phrases and their compositionality.
\newblock In \emph{Advances in Neural Information Processing Systems 26}, pages 3111--3119. Curran Associates, Inc.

\bibitem[Mikolov et~al.(2013{\natexlab{c}})Mikolov, Yih, and Zweig]{mikolov-etal-2013-linguistic}
{\rm Mikolov, T., Yih, W.-t., {\rm and} Zweig, G.} (2013{\natexlab{c}}).
\newblock Linguistic regularities in continuous space word representations.
\newblock In \emph{Proceedings of the 2013 {C}onference of the {N}orth {A}merican {C}hapter of the {A}ssociation for {C}omputational {L}inguistics: {H}uman {L}anguage {T}echnologies}, pages 746--751.

\bibitem[Mitra et~al.(2014)Mitra, Mitra, Riedl, Biemann, Mukherjee, and Goyal]{2014arXiv1405.4392M}
{\rm Mitra, S., Mitra, R., Riedl, M., Biemann, C., Mukherjee, A., {\rm and} Goyal, P.} (2014).
\newblock {That's sick dude!: Automatic identification of word sense change across different timescales}.
\newblock \emph{arXiv e-prints}, art. arXiv:1405.4392.

\bibitem[Mitra et~al.(2015)Mitra, Mitra, Maity, Riedl, Biemann, Goyal, and Mukherjee]{mitra2015automatic}
{\rm Mitra, S., Mitra, R., Maity, S.~K., Riedl, M., Biemann, C., Goyal, P., {\rm and} Mukherjee, A.} (2015).
\newblock An automatic approach to identify word sense changes in text media across timescales.
\newblock \emph{Natural Language Engineering}, {\bf 21}\penalty0 (5), \penalty0 773--798.

\bibitem[{Montanelli} and {Periti}(2023)]{2023arXiv230401666M}
{\rm {Montanelli}, S. {\rm and} {Periti}, F.} (2023).
\newblock {A Survey on Contextualised Semantic Shift Detection}.
\newblock \emph{arXiv e-prints}, art. arXiv:2304.01666.
\newblock \doi{10.48550/arXiv.2304.01666}.

\bibitem[Naseem et~al.(2021)Naseem, Razzak, Khan, and Prasad]{10.1145/3434237}
{\rm Naseem, U., Razzak, I., Khan, S.~K., {\rm and} Prasad, M.} (2021).
\newblock A comprehensive survey on word representation models: From classical to state-of-the-art word representation language models.
\newblock \emph{ACM Trans. Asian Low-Resour. Lang. Inf. Process.}, {\bf 20}\penalty0 (5).
\newblock ISSN 2375-4699.
\newblock \doi{10.1145/3434237}.
\newblock URL \url{https://doi.org/10.1145/3434237}.

\bibitem[{Neal}(2012)]{2012arXiv1206.1901N}
{\rm {Neal}, R.~M.} (2012).
\newblock {MCMC using {H}amiltonian dynamics}.
\newblock \emph{arXiv e-prints}, art. arXiv:1206.1901.

\bibitem[Patel and Bhattacharyya(2017)]{patel-bhattacharyya-2017-towards}
{\rm Patel, K. {\rm and} Bhattacharyya, P.} (2017).
\newblock Towards lower bounds on number of dimensions for word embeddings.
\newblock In \emph{Proceedings of the Eighth International Joint Conference on Natural Language Processing (Volume 2: Short Papers)}, pages 31--36, Taipei, Taiwan, November 2017. Asian Federation of Natural Language Processing.
\newblock URL \url{https://aclanthology.org/I17-2006}.

\bibitem[Pennington et~al.(2014)Pennington, Socher, and Manning]{pennington2014glove}
{\rm Pennington, J., Socher, R., {\rm and} Manning, C.~D.} (2014).
\newblock Glove: Global vectors for word representation.
\newblock In \emph{Empirical Methods in Natural Language Processing (EMNLP)}, pages 1532--1543.

\bibitem[Perrone et~al.(2019)Perrone, Palma, Hengchen, Vatri, Smith, and McGillivray]{perrone-etal-2019-gasc}
{\rm Perrone, V., Palma, M., Hengchen, S., Vatri, A., Smith, J.~Q., {\rm and} McGillivray, B.} (2019).
\newblock {GASC}: Genre-aware semantic change for {A}ncient {G}reek.
\newblock In \emph{Proceedings of the 1st International Workshop on Computational Approaches to Historical Language Change}, pages 56--66, Florence, Italy, August 2019. Association for Computational Linguistics.
\newblock \doi{10.18653/v1/W19-4707}.

\bibitem[Peters et~al.(2018)Peters, Neumann, Iyyer, Gardner, Clark, Lee, and Zettlemoyer]{peters-etal-2018-deep}
{\rm Peters, M.~E., Neumann, M., Iyyer, M., Gardner, M., Clark, C., Lee, K., {\rm and} Zettlemoyer, L.} (2018).
\newblock Deep contextualized word representations.
\newblock In \emph{Proceedings of the 2018 Conference of the North {A}merican Chapter of the Association for Computational Linguistics: Human Language Technologies, Volume 1 (Long Papers)}, pages 2227--2237, New Orleans, Louisiana, June 2018. Association for Computational Linguistics.
\newblock \doi{10.18653/v1/N18-1202}.
\newblock URL \url{https://aclanthology.org/N18-1202}.

\bibitem[{R Core Team}(2023)]{R_citation}
{\rm {R Core Team}} (2023).
\newblock \emph{R: A Language and Environment for Statistical Computing}.
\newblock R Foundation for Statistical Computing, Vienna, Austria.
\newblock URL \url{https://www.R-project.org/}.

\bibitem[Radford et~al.(2019)Radford, Wu, Child, Luan, Amodei, Sutskever, et~al.]{radford2019language}
{\rm Radford, A., Wu, J., Child, R., Luan, D., Amodei, D., Sutskever, I., et~al.} (2019).
\newblock Language models are unsupervised multitask learners.
\newblock \emph{OpenAI blog}, {\bf 1}\penalty0 (8), \penalty0 9.

\bibitem[Roberts and Rosenthal(1998)]{doi:10.1111/1467-9868.00123}
{\rm Roberts, G.~O. {\rm and} Rosenthal, J.~S.} (1998).
\newblock Optimal scaling of discrete approximations to {L}angevin diffusions.
\newblock \emph{Journal of the Royal Statistical Society: Series B}, {\bf 60}\penalty0 (1), \penalty0 255--268.
\newblock \doi{10.1111/1467-9868.00123}.

\bibitem[Roberts and Tweedie(1996)]{10.2307/3318418}
{\rm Roberts, G.~O. {\rm and} Tweedie, R.~L.} (1996).
\newblock Exponential convergence of {L}angevin distributions and their discrete approximations.
\newblock \emph{Bernoulli}, {\bf 2}\penalty0 (4), \penalty0 341--363.
\newblock ISSN 13507265.

\bibitem[Rodda et~al.(2019)Rodda, Probert, and McGillivray]{rodda2019a}
{\rm Rodda, M., Probert, P., {\rm and} McGillivray, B.} (2019).
\newblock Vector space models of {A}ncient {G}reek word meaning, and a case study on {H}omer.
\newblock \emph{Traitement Automatique des Langues}, {\bf 60}\penalty0 (3), \penalty0 63--87.

\bibitem[Rudolph and Blei(2018)]{Rudolph:2018:DEL:3178876.3185999}
{\rm Rudolph, M. {\rm and} Blei, D.} (2018).
\newblock Dynamic embeddings for language evolution.
\newblock In \emph{Proceedings of the 2018 World Wide Web Conference}, pages 1003--1011.

\bibitem[Selivanov(2022)]{GloVe_code}
{\rm Selivanov, D.} (2022).
\newblock Glove word embeddings.
\newblock URL \url{https://cran.r-project.org/web/packages/text2vec/vignettes/glove.html}.
\newblock Accessed 2022-06-01.

\bibitem[Shaby and Wells(2010)]{shaby2010exploring}
{\rm Shaby, B. {\rm and} Wells, M.~T.} (2010).
\newblock Exploring an adaptive {M}etropolis algorithm.

\bibitem[{Stan Development Team}(2023{\natexlab{a}})]{RStan_citation}
{\rm {Stan Development Team}} (2023{\natexlab{a}}).
\newblock {RStan}: the {R} interface to {Stan}.
\newblock URL \url{https://mc-stan.org/}.
\newblock R package version 2.32.3.

\bibitem[{Stan Development Team}(2023{\natexlab{b}})]{Stan_citation}
{\rm {Stan Development Team}} (2023{\natexlab{b}}).
\newblock Stan modeling language user's guide and reference manual.
\newblock URL \url{https://mc-stan.org/docs/2_26/reference-manual/index.html}.
\newblock Stan version 2.26.1.

\bibitem[{Statisticat} and {LLC.}(2021)]{R_LaplacesDemon}
{\rm {Statisticat} {\rm and} {LLC.}} (2021).
\newblock \emph{LaplacesDemon: Complete Environment for {B}ayesian Inference}.
\newblock URL \url{https://web.archive.org/web/20150206004624/http://www.{B}ayesian-inference.com/software}.
\newblock R package version 16.1.6.

\bibitem[Syed et~al.(2021)Syed, Bouchard-Côté, Deligiannidis, and Doucet]{10.1111/rssb.12464}
{\rm Syed, S., Bouchard-Côté, A., Deligiannidis, G., {\rm and} Doucet, A.} (2021).
\newblock {Non-Reversible Parallel Tempering: A Scalable Highly Parallel MCMC Scheme}.
\newblock \emph{Journal of the Royal Statistical Society Series B: Statistical Methodology}, {\bf 84}\penalty0 (2), \penalty0 321--350.
\newblock ISSN 1369-7412.
\newblock \doi{10.1111/rssb.12464}.
\newblock URL \url{https://doi.org/10.1111/rssb.12464}.

\bibitem[Tahmasebi and Risse(2017)]{tahmasebi2017finding}
{\rm Tahmasebi, N. {\rm and} Risse, T.} (2017).
\newblock Finding individual word sense changes and their delay in appearance.
\newblock In \emph{RANLP}, pages 741--749.

\bibitem[Tahmasebi et~al.(2021)Tahmasebi, Borin, and Jatowt]{tahmasebi2021survey}
{\rm Tahmasebi, N., Borin, L., {\rm and} Jatowt, A.} (2021).
\newblock Survey of computational approaches to lexical semantic change detection.
\newblock \emph{Computational approaches to semantic change}, pages 1--91.

\bibitem[Tang(2018)]{tang2018state}
{\rm Tang, X.} (2018).
\newblock A state-of-the-art of semantic change computation.
\newblock \emph{Natural Language Engineering}, {\bf 24}\penalty0 (5).

\bibitem[Vatri and McGillivray(2018)]{TheDiorisisAncientGreekCorpus}
{\rm Vatri, A. {\rm and} McGillivray, B.} (2018).
\newblock The {D}iorisis {A}ncient {G}reek corpus.
\newblock \emph{Research Data Journal for the Humanities and Social Sciences}, {\bf 3}\penalty0 (1), \penalty0 55 -- 65.

\bibitem[Vatri et~al.(2019)Vatri, Lähteenoja, and McGillivray]{vatri_lahteenoja_mcgillivray_2019}
{\rm Vatri, A., Lähteenoja, V., {\rm and} McGillivray, B.} (2019).
\newblock {A}ncient {G}reek semantic change - annotated datasets and code.

\bibitem[Vehtari et~al.(2017)Vehtari, Gelman, and Gabry]{vehtari2017practical}
{\rm Vehtari, A., Gelman, A., {\rm and} Gabry, J.} (2017).
\newblock Practical {B}ayesian model evaluation using leave-one-out cross-validation and waic.
\newblock \emph{Statistics and computing}, {\bf 27}\penalty0 (5), \penalty0 1413--1432.

\bibitem[Vehtari et~al.(2024)Vehtari, Gabry, Magnusson, Yao, Bürkner, Paananen, and Gelman]{R_loo}
{\rm Vehtari, A., Gabry, J., Magnusson, M., Yao, Y., Bürkner, P.-C., Paananen, T., {\rm and} Gelman, A.} (2024).
\newblock loo: Efficient leave-one-out cross-validation and waic for {B}ayesian models.
\newblock URL \url{https://mc-stan.org/loo/}.
\newblock R package version 2.7.0.

\bibitem[Watanabe and Opper(2010)]{watanabe2010asymptotic}
{\rm Watanabe, S. {\rm and} Opper, M.} (2010).
\newblock Asymptotic equivalence of {B}ayes cross validation and widely applicable information criterion in singular learning theory.
\newblock \emph{Journal of machine learning research}, {\bf 11}\penalty0 (12).

\bibitem[Yin and Shen(2018)]{yin2018dimensionality}
{\rm Yin, Z. {\rm and} Shen, Y.} (2018).
\newblock On the dimensionality of word embedding.
\newblock \emph{Advances in neural information processing systems}, {\bf 31}.

\bibitem[Yüksel et~al.(2021)Yüksel, Uğurlu, and Koç]{9581306}
{\rm Yüksel, A., Uğurlu, B., {\rm and} Koç, A.} (2021).
\newblock Semantic change detection with gaussian word embeddings.
\newblock \emph{IEEE/ACM Transactions on Audio, Speech, and Language Processing}, {\bf 29}, \penalty0 3349--3361.
\newblock \doi{10.1109/TASLP.2021.3120645}.

\bibitem[Zafar and Nicholls(2022)]{DiSC_https://doi.org/10.1111/rssc.12591}
{\rm Zafar, S. {\rm and} Nicholls, G.~K.} (2022).
\newblock Measuring diachronic sense change: {N}ew models and {M}onte {C}arlo methods for {B}ayesian inference.
\newblock \emph{Journal of the Royal Statistical Society: Series C (Applied Statistics)}, {\bf 71}\penalty0 (5), \penalty0 1569--1604.
\newblock \doi{https://doi.org/10.1111/rssc.12591}.

\end{thebibliography}

\end{document}